\documentclass[10pt,twocolumn,letterpaper]{article}

\usepackage{cvpr}
\usepackage{times}
\usepackage{epsfig}
\usepackage{graphicx}
\usepackage{amsmath}
\usepackage{amssymb}
\usepackage{float}
\usepackage{booktabs}
\usepackage{nicefrac}
\usepackage[colorlinks,linkcolor=red]{hyperref}

\cvprfinalcopy % *** Uncomment this line for the final submission

\def\cvprPaperID{963} % *** Enter the CVPR Paper ID here

\ifcvprfinal\fi
\begin{document}

%%%%%%%%% TITLE
\title{Pluralistic Image Completion}

\author{Chuanxia Zheng \qquad Tat-Jen Cham \qquad Jianfei Cai\\
	School of Computer Science and Engineering \\
	Nanyang Technological University, Singapore\\
	{\tt\small \{chuanxia001,astjcham,asjfcai\}@ntu.edu.sg}
}

\twocolumn[{%
	\maketitle
	\vspace*{-1.0cm}
	\begin{figure}[H]
		\centering
		\hsize=\textwidth % cvpr 需要
		\includegraphics[width=\textwidth]{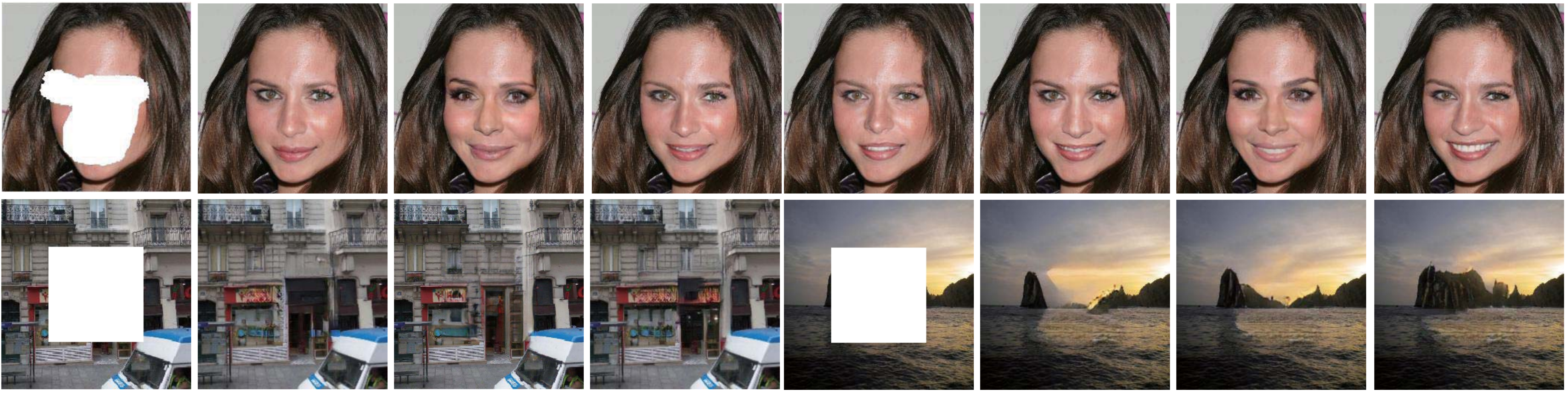}
		\caption{Example completion results of our method on images of a face, a building, and natural scenery with various masks (missing regions shown in white). For each group, the masked input image is shown left, followed by sampled results from our model without any post-processing. The results are diverse and plausible. (Zoom in to see the details.)}
		\label{fig:example}
	\end{figure}
}]
%\maketitle
%\thispagestyle{empty}

%%%%%%%%% ABSTRACT
\begin{abstract}
Most image completion methods produce only one result for each masked input, although there may be many reasonable possibilities. In this paper, we present an approach for \textbf{pluralistic image completion} -- the task of generating multiple and diverse plausible solutions for image completion. A major challenge faced by learning-based approaches is that usually only one ground truth training instance per label. As such, sampling from conditional VAEs still leads to minimal diversity. To overcome this, we propose a novel and probabilistically principled framework with two parallel paths. One is a reconstructive path that utilizes the only one given ground truth to get prior distribution of missing parts and rebuild the original image from this distribution. The other is a generative path for which the conditional prior is coupled to the distribution obtained in the reconstructive path. Both are supported by GANs. We also introduce a new short+long term attention layer that exploits distant relations among decoder and encoder features, improving appearance consistency. When tested on datasets with buildings (Paris), faces (CelebA-HQ), and natural images (ImageNet), our method not only generated higher-quality completion results, but also with multiple and diverse plausible outputs.
\end{abstract}

%%%%%%%%% BODY TEXT
\section{Introduction}

Image completion is a highly subjective process. Supposing you were shown the various images with missing regions in fig. \ref{fig:example}, what would you \emph{imagine} to be occupying these holes? Bertalmio \etal \cite{bertalmio2000image} related how expert conservators would inpaint damaged art by: 1) imagining the semantic content to be filled based on the overall scene; 2) ensuring structural continuity between the masked and unmasked regions; and 3) filling in visually realistic content for missing regions. Nonetheless, each expert will independently end up creating substantially different details, even if they may universally agree on high-level semantics, such as general placement of eyes on a damaged portrait. 

Based on this observation, our main goal is thus to generate \emph{multiple} and \emph{diverse} plausible results when presented with a masked image --- in this paper we refer to this task as \textbf{pluralistic image completion} (depicted in fig. \ref{fig:example}). This is as opposed to approaches that attempt to generate only a single ``guess'' for missing parts.  

%Instead of fixating on recreating the original image in an image completion task, we embrace the large variety of reasonable depictions as part of an intrinsically creative process. With this tenet in mind

Early image completion works \cite{bertalmio2000image, criminisi2003object, bertalmio2003simultaneous, criminisi2004region, barnes2009patchmatch, huang2014image} focus only on steps 2 and 3 above, by assuming that gaps should be filled with similar content to that of the background. Although these approaches produced high-quality texture-consistent images, they cannot capture global semantics and hallucinate new content for large holes. More recently, some learning-based image completion methods \cite{pathak2016context, iizuka2017globally, yang2017high, yeh2017semantic, yu2018generative, Liu_2018_ECCV, Yan_2018_ECCV} were proposed that infer semantic content (as in step 1). These works treated completion as a conditional generation problem, where the input-to-output mapping is one-to-many. However, these prior works are limited to generate only one ``optimal'' result, and do not have the capacity to generate a variety of semantically meaningful results. 

To obtain a diverse set of results, some methods utilize conditional variational auto-encoders (CVAE)~\cite{sohn2015learning, walker2016, bao2017cvae, Eslami2018}, a conditional extension of  VAE~\cite{kingma2013auto}, which explicitly code a distribution that can be sampled. However, specifically for an image completion scenario, the standard single-path formulation usually leads to grossly underestimating variances. This is because when \emph{the condition label is itself a partial image}, the number of instances in the training data that match each label is \emph{typically only one}. Hence the estimated conditional distributions tend to have very limited variation since they were trained to reconstruct the single ground truth. This is further elaborated on in section~\ref{framework}.

An important insight we will use is that \emph{partial images}, as a superset of full images, may also be considered as generated from \emph{a latent space with smooth prior distributions}. This provides a mechanism for alleviating the problem of having scarce samples per conditional partial image. To do so, we introduce a new image completion network with two parallel but linked training pipelines. The first pipeline is a VAE-based reconstructive path that not only utilizes the full instance ground truth (\ie both the visible partial image, as well as its complement --- the hidden partial image), but also imposes smooth priors for the latent space of complement regions. The second pipeline is a generative path that predicts the latent prior distribution for the missing regions conditioned on the visible pixels, from which can be sampled to generate diverse results. The training process for the latter path does \emph{not} attempt to steer the output towards reconstructing the instance-specific hidden pixels at all, instead allowing the reasonableness of results be driven by an auxiliary discriminator network \cite{goodfellow2014generative}. This leads to substantially great variability in content generation. We also introduce an enhanced short+long term attention layer that significantly increases the quality of our results. 

We compared our method with existing state-of-the-art approaches on multiple datasets. Not only can higher-quality completion results be generated using our approach, it also presents multiple diverse solutions.

The main contributions of this work are:
\begin{enumerate}
	\item A probabilistically principled framework for image completion that is able to maintain much higher sample diversity as compared to existing methods;
	\item A new network structure with two parallel training paths, which trades off between reconstructing the original training data (with loss of diversity) and maintaining the variance of the conditional distribution;
	\item A novel self-attention layer that exploits short+long term context information to ensure appearance consistency in the image domain, in a manner superior to purely using GANs; and
	\item We demonstrate that our method is able to complete the same mask with multiple plausible results that have substantial diversity, such as those shown in figure~\ref{fig:example}. 
\end{enumerate}

%-------------------------------------------------------------------------
\section{Related Work}

Existing work on image completion either uses information from within the input image  \cite{bertalmio2000image, bertalmio2003simultaneous, barnes2009patchmatch}, or information from a large image dataset \cite{hays2007scene, pathak2016context, yu2018generative}. Most approaches will generate only one result per masked image.

\noindent\textbf{Intra-Image Completion} Traditional intra-image completion, such as diffusion-based methods \cite{bertalmio2000image, ballester2001filling, levin2003learning} and patch-based methods \cite{bertalmio2003simultaneous, criminisi2003object, criminisi2004region, barnes2009patchmatch}, assume image holes share similar content to visible regions; thus they would directly match, copy and realign the background patches to complete the holes. These methods perform well for background completion, \eg for object removal, but cannot hallucinate unique content not present in the input images.  

\noindent\textbf{Inter-Image Completion} To generate semantically new content, inter-image completion borrows information from a large dataset. Hays and Efros~\cite{hays2007scene} presented an image completion method using millions of images, in which the image most similar to the masked input is retrieved, and corresponding regions are transferred. However, this requires a high contextual match, which is not always available. Recently, learning-based approaches were proposed. Initial works \cite{kohler2014mask, ren2015shepard} focused on small and thin holes. Context encoders (CE) \cite{pathak2016context} handled 64$\times$64-sized holes using GANs \cite{goodfellow2014generative}. This was followed by several CNN-based methods, which included combining global and local discriminators as adversarial loss \cite{iizuka2017globally}, identifying closest features in the latent space of masked images \cite{yeh2017semantic}, utilizing semantic labels to guide the completion network \cite{song2018spg}, introducing additional face parsing loss for face completion \cite{li2017generative}, and designing particular convolutions to address irregular holes \cite{Liu_2018_ECCV, yu2018free}. A common drawback of these methods is that they often create distorted structures and blurry textures inconsistent with the visible regions, especially for large holes. 

\noindent\textbf{Combined Intra- and Inter-Image Completion} To overcome the above problems, Yang \etal\cite{yang2017high} proposed multi-scale neural patch synthesis, which generates high-frequency details by copying patches from mid-layer features. However, this optimization is computational costly. More recently, several works \cite{yu2018generative, Yan_2018_ECCV, song2018contextual} exploited spatial attention \cite{jaderberg2015spatial, zhou2016view} to get high-frequency details. Yu \etal\cite{yu2018generative} presented a contextual attention layer to copy similar features from visible regions to the holes. Yan \etal\cite{Yan_2018_ECCV} and Song \etal\cite{song2018contextual} proposed PatchMatch-like ideas on feature domain. However, these methods identify similar features by comparing features of holes and features of visible regions, which is somewhat contradictory as feature transfer is unnecessary when two features are very similar, but when needed the features are too different to be matched easily. Furthermore, distant information is not used for new content that differs from visible regions. Our model will solve this problem by extending self-attention \cite{zhang2018self} to harness abundant context.

\noindent\textbf{Image Generation} Image generation has progressed significantly using methods such as VAE \cite{kingma2013auto} and GANs \cite{goodfellow2014generative}. These have been applied to conditional image generation tasks, such as image translation \cite{isola2017image}, synthetic to realistic \cite{zheng2018t2net}, future prediction \cite{mathieu2015deep}, and 3D models \cite{park2017transformation}. Perhaps most relevant are conditional VAEs (CVAE) \cite{sohn2015learning,walker2016} and CVAE-GAN \cite{bao2017cvae}, but these were not specially targeted for image completion. CVAE-based methods are most useful when the conditional labels are few and discrete, and there are sufficient training instances per label. Some recent work utilizing these in image translation can produce diverse output~\cite{zhu2017toward,lee2018diverse}, but in such situations the condition-to-sample mappings are more local (\eg pixel-to-pixel), and only change the visual appearance. This is untrue for image completion, where the conditional label is itself the masked image, with only one training instance of the original holes. In \cite{chen2018high}, different outputs were obtained for face completion by specifying facial attributes (\eg smile), but this method is very domain specific, requiring targeted attributes.

\section{Approach}

\begin{figure}[tb!]
	\centering
	\includegraphics[width=0.97\linewidth]{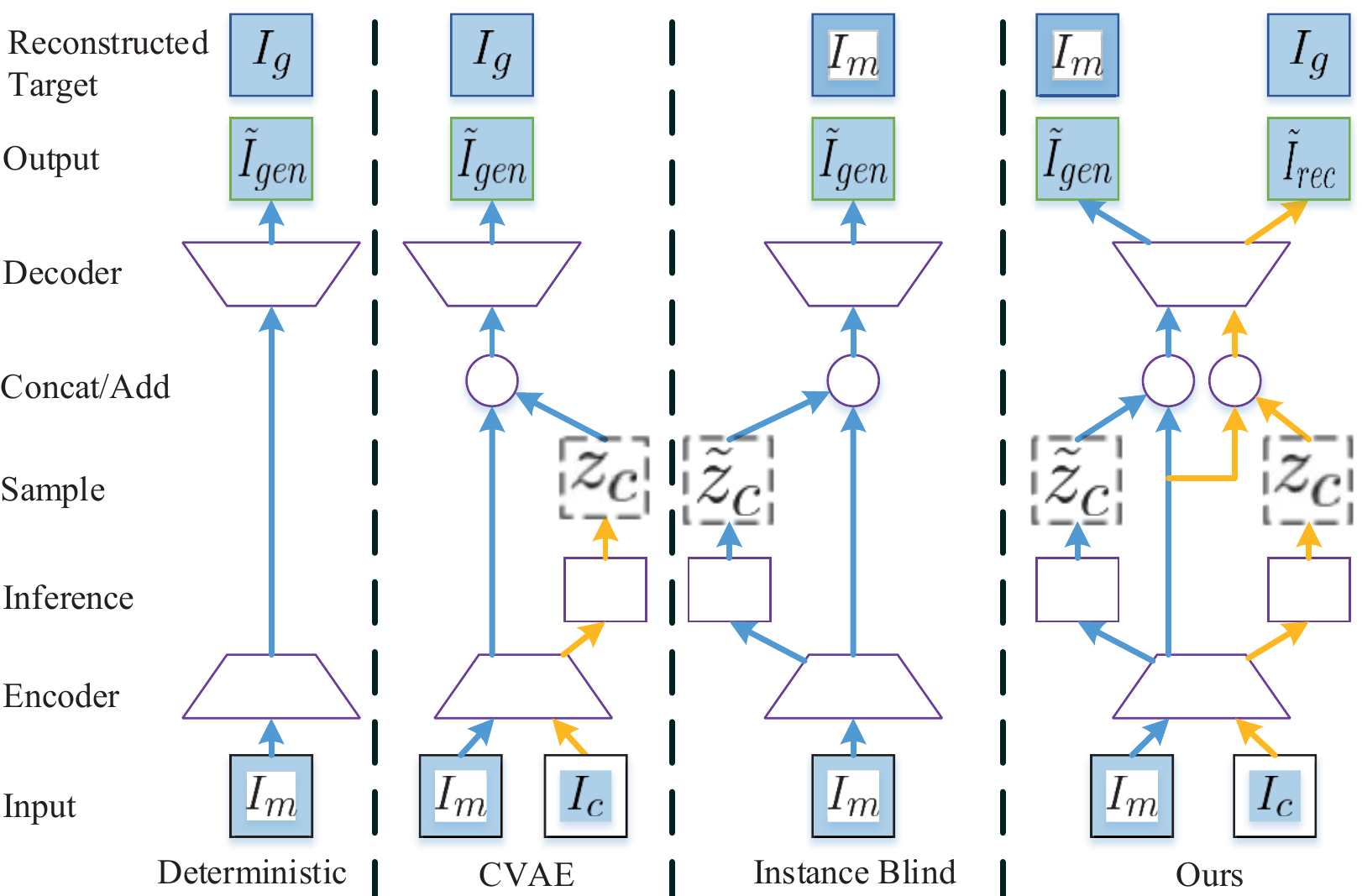}
	\caption{Completion strategies given masked input. (Deterministic) structure directly predicts the ground truth instance. (CVAE) adds in random sampling to diversify the output. (Instance Blind) only matches the visible parts, but training is unstable. (Ours) uses a generative path during testing, but is guided by a parallel reconstructive path during training. Yellow path is used for training.}
	\label{fig:coarse_framework}	
\end{figure}

Suppose we have an image, originally $\mathbf{I}_g$, but degraded by a number of missing pixels to become $\mathbf{I}_m$ (the \emph{masked partial image}) comprising the observed / visible pixels. We also define $\mathbf{I}_c$ as its \emph{complement partial image} comprising the ground truth hidden pixels. Classical image completion methods attempt to reconstruct the ground truth unmasked image $\mathbf{I}_g$ in a deterministic fashion from $\mathbf{I}_m$ (see fig.~\ref{fig:coarse_framework} ``Deterministic''). This results in only a single solution. In contrast, our goal is to \emph{sample} from $p(\mathbf{I}_c|\mathbf{I}_m)$. 

\begin{figure*}[tb!]
	\centering
	\includegraphics[width=\textwidth,height=0.30\textheight]{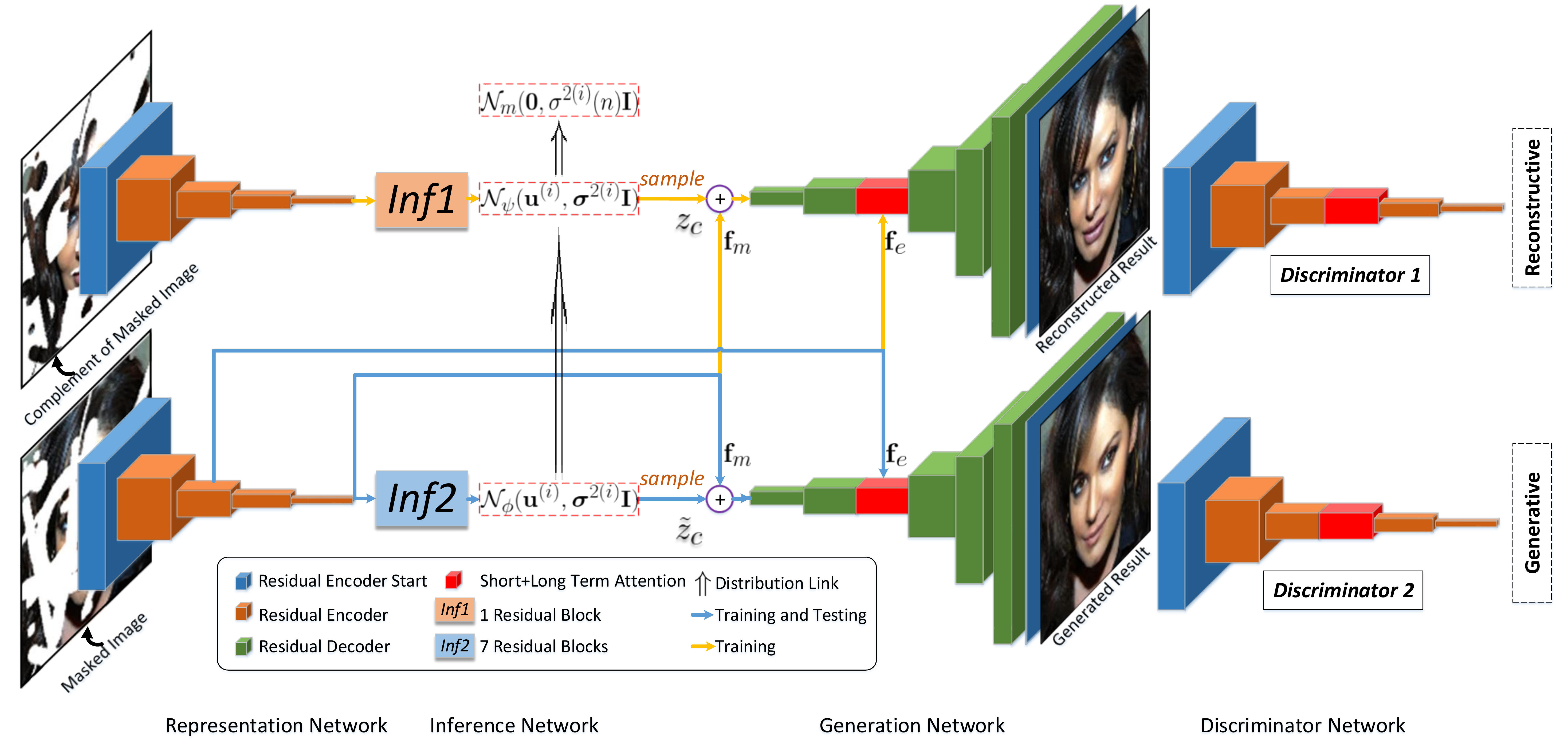}
	\caption{Overview of our architecture with two parallel pipelines. The \textbf{reconstructive} pipeline (yellow line) combines information from $\mathbf{I}_m$ and $\mathbf{I}_c$, which is used only for training. The \textbf{generative} pipeline (blue line) infers the conditional distribution of hidden regions, that can be sampled during testing. Both representation and generation networks share identical weights.}
	\label{fig:framework}	
\end{figure*}

\subsection{Probabilistic Framework}\label{framework}

In order to have a distribution to sample from, a current approach is to employ the CVAE~\cite{sohn2015learning} which estimates a parametric distribution over a latent space, from which sampling is possible (see fig.~\ref{fig:coarse_framework} ``CVAE''). This involves a variational lower bound of the conditional log-likelihood of observing the training instances:
\begin{align}\label{eq:CVAE}
\log p(\mathbf{I}_c|\mathbf{I}_m)\ge& -\text{KL}(q_\psi(\mathbf{z}_c|\mathbf{I}_c,\mathbf{I}_m)||p_\phi(\mathbf{z}_c|\mathbf{I}_m)) \nonumber\\ & +\mathbb{E}_{q_\psi(\mathbf{z}_c|\mathbf{I}_c,\mathbf{I}_m)}[\log p_\theta(\mathbf{I}_c|\mathbf{z}_c,\mathbf{I}_m)]
\end{align}
where $\mathbf{z}_c$ is the latent vector,  $q_\psi(\cdot|\cdot)$ the posterior importance sampling function, $p_\phi(\cdot|\cdot)$ the conditional prior, $p_\theta(\cdot|\cdot)$ the likelihood, with $\psi$, $\phi$ and $\theta$ being the deep network parameters of their corresponding functions. This lower bound is maximized \wrt all parameters.

For our purposes, the chief difficulty of using CVAE \cite{sohn2015learning} directly is that the high DoF networks of $q_\psi(\cdot|\cdot)$ and $p_\phi(\cdot|\cdot)$ are not easily separable in (\ref{eq:CVAE}) with the KL distance easily driven towards zero, and is approximately equivalent to maximizing $\mathbb{E}_{p_\phi(\mathbf{z}_c|\mathbf{I}_m)}[\log p_\theta(\mathbf{I}_c|\mathbf{z}_c,\mathbf{I}_m)]$ (the ``GSNN'' variant in \cite{sohn2015learning}). This consequently learns a delta-like prior of $p_\phi(\mathbf{z}_c|\mathbf{I}_m)\rightarrow \delta(\mathbf{z}_c-\mathbf{z}^*_c)$, where $\mathbf{z}^*_c$ is the maximum latent likelihood point of $p_\theta(\mathbf{I}_c|\cdot,\mathbf{I}_m)$. While this low variance prior may be useful in estimating a single solution, sampling from it will lead to \emph{negligible diversity} in image completion results (as seen in fig.~\ref{fig:ablation_paths}). When the CVAE variant of \cite{walker2016}, which has a fixed latent prior, is used instead, the network learns to ignore the latent sampling and directly estimates $\mathbf{I}_c$ from $\mathbf{I}_m$, also resulting in a single solution.  This is due to the image completion scenario when there is only one training instance per condition label, which is a partial image $\mathbf{I}_m$. Details are in the supplemental section~\ref{appsec:difficulties}.

A possible way to diversify the output is to simply not incentivize the output to reconstruct the instance-specific $\mathbf{I}_g$ during training, only needing it to fit in with the training set distribution as deemed by an  learned adversarial discriminator (see fig.~\ref{fig:coarse_framework} ``Instance Blind''). However, this approach is unstable, especially for large and complex scenes~\cite{song2018contextual}.

\noindent\textbf{Latent Priors of Holes}
In our approach, we require that missing partial images, as a superset of full images, \emph{to also arise from a latent space distribution}, with a smooth prior of $p(\mathbf{z}_c)$. The variational lower bound is:
\begin{align}\label{eq:VAE}
\log p(\mathbf{I}_c)\ge& -\text{KL}(q_\psi(\mathbf{z}_c|\mathbf{I}_c)||p(\mathbf{z}_c)) \nonumber\\ & +\mathbb{E}_{q_\psi(\mathbf{z}_c|\mathbf{I}_c)}[\log p_\theta(\mathbf{I}_c|\mathbf{z}_c)]
\end{align}
where in \cite{kingma2013auto} the prior is set as $p(\mathbf{z}_c)=\mathcal{N}(\mathbf{0},\mathbf{I})$. However, we can be more discerning when it comes to partial images since they have different numbers of pixels. \emph{A missing partial image $\mathbf{z}_c$ with more pixels (larger holes) should have greater latent prior variance than a missing partial image $\mathbf{z}_c$ with fewer pixels (smaller holes)}. Hence we generalize the prior $p(\mathbf{z}_c)=\mathcal{N}_m(\mathbf{0},\sigma^2(n)\mathbf{I})$ to adapt to the number of pixels $n$. 

%, and in fact a missing partial image with no pixels (no holes) should be completely deterministic!

\paragraph{Prior-Conditional Coupling}
Next, we combine the latent priors into the conditional lower bound of (\ref{eq:CVAE}). This can be done by assuming $\mathbf{z}_c$ is much more closely related to $\mathbf{I}_c$ than to $\mathbf{I}_m$, so $q_\psi(\mathbf{z}_c|\mathbf{I}_c,\mathbf{I}_m)$$\approx$$q_\psi(\mathbf{z}_c|\mathbf{I}_c)$. Updating (\ref{eq:CVAE}):
\begin{align}\label{eq:CVAE_with_prior}
\log p(\mathbf{I}_c|\mathbf{I}_m)\ge& -\text{KL}(q_\psi(\mathbf{z}_c|\mathbf{I}_c)||p_\phi(\mathbf{z}_c|\mathbf{I}_m)) \nonumber\\ & +\mathbb{E}_{q_\psi(\mathbf{z}_c|\mathbf{I}_c)}[\log p_\theta(\mathbf{I}_c|\mathbf{z}_c,\mathbf{I}_m)]
\end{align}
However, unlike in (\ref{eq:CVAE}), notice that $q_\psi(\mathbf{z}_c|\mathbf{I}_c)$ \emph{is no longer freely learned during training, but is tied to its presence in (\ref{eq:VAE})}. Intuitively, the learning of $q_\psi(\mathbf{z}_c|\mathbf{I}_c)$ is regularized by the prior $p(\mathbf{z}_c)$ in (\ref{eq:VAE}), while the learning of the conditional prior $p_\phi(\mathbf{z}_c|\mathbf{I}_m)$ is in turn regularized by $q_\psi(\mathbf{z}_c|\mathbf{I}_c)$ in (\ref{eq:CVAE_with_prior}).

\paragraph{Reconstruction vs Creative Generation}
\label{sec:rec_vs_gen}
One issue with (\ref{eq:CVAE_with_prior}) is that the sampling is taken from $q_\psi(\mathbf{z}_c|\mathbf{I}_c)$ during training, but is not available during testing, whereupon sampling must come from $p_\phi(\mathbf{z}_c|\mathbf{I}_m)$ which may not be adequately learned for this role. In order to mitigate this problem, we modify (\ref{eq:CVAE_with_prior}) to have a blend of formulations \emph{with and without importance sampling}. So, with simplified notation:
\begin{align}\label{eq:mixed_models}
\log p(\mathbf{I}_c|\mathbf{I}_m) \geq &
\lambda \left\{ \mathbb{E}_{q_\psi}[\log p^r_\theta(\mathbf{I}_c|\mathbf{z}_c,\mathbf{I}_m)]
-  \text{KL}(q_\psi || p_\phi) \right\}\nonumber\\
& + (1-\lambda)\,  \mathbb{E}_{p_\phi}[
\log p^g_\theta(\mathbf{I}_c|\mathbf{z}_c,\mathbf{I}_m)]
\end{align}
where $0\leq \lambda \leq 1$ is implicitly set by training loss coefficients in section~\ref{sec:training_loss}. When sampling from the importance function $q_\psi(\cdot|\mathbf{I}_c)$, the full training instance is available and we formulate the likelihood $p^r_\theta(\mathbf{I}_c|\mathbf{z}_c,\mathbf{I}_m)$ to be focused on \emph{reconstructing} $\mathbf{I}_c$. Conversely, when sampling from the learned conditional prior $p_\phi(\cdot|\mathbf{I}_m)$ which does not contain $\mathbf{I}_c$, we facilitate \emph{creative generation} by having the likelihood model $p^g_\theta(\mathbf{I}_c|\mathbf{z}_c,\mathbf{I}_m) \cong \ell^g_\theta(\mathbf{z}_c,\mathbf{I}_m)$ be \emph{independent of the original instance} of $\mathbf{I}_c$. Instead it only encourages generated samples to fit in with the overall training distribution.

Our overall training objective may then be expressed as jointly maximizing the lower bounds in (\ref{eq:VAE}) and (\ref{eq:mixed_models}), with the likelihood in (\ref{eq:VAE}) unified to that in (\ref{eq:mixed_models}) as $p_\theta(\mathbf{I}_c|\mathbf{z}_c)\cong p^r_\theta(\mathbf{I}_c|\mathbf{z}_c,\mathbf{I}_m)$. See the supplemental section~\ref{appsec:joint_maximization}.

\subsection{Dual Pipeline Network Structure}
\label{sec:dual_pipeline_structure}
This formulation is implemented as our dual pipeline framework, shown in fig.~\ref{fig:framework}. It consists of two paths: the upper reconstructive path uses information from the whole image, \ie $\mathbf{I}_g$=$\{\mathbf{I}_c,\mathbf{I}_m\}$, while the lower generative path only uses information from visible regions $\mathbf{I}_m$. Both representation and generation networks share identical weights. Specifically:
\begin{itemize}
	\item For the upper reconstructive path, the complement partial image $\mathbf{I}_c$ is used to infer the importance function $q_\psi(\cdot|\mathbf{I}_c)$=$\mathcal{N}_\psi(\cdot)$ during training. The sampled latent vector $\mathbf{z}_c$ thus contains information of the missing regions, while the conditional feature $\mathbf{f}_m$ encodes the information of the visible regions. Since there is sufficient information, the loss function in this path is geared towards reconstructing the original image $\mathbf{I}_g$.
	\item For the lower generative path, which is also the test path, the latent distribution of the holes $\mathbf{I}_c$ is inferred based only on the visible $\mathbf{I}_m$. This would be significantly less accurate than the inference in the upper path. Thus the reconstruction loss is only targeted at the visible regions $\mathbf{I}_m$ (via $\mathbf{f}_m$).
	\item In addition, we also utilize adversarial learning networks on both paths, which ideally ensure that the full synthesized data fit in with the training set distribution, and empirically leads to higher quality images. 
\end{itemize}

\subsection{Training Loss}
\label{sec:training_loss}

Various terms in (\ref{eq:VAE}) and (\ref{eq:mixed_models}) may be more conventionally expressed as loss functions. Jointly maximizing the lower bounds is then minimizing a total loss $\mathcal{L}$, which consists of three groups of component losses:
\begin{equation}
\mathcal{L} = \alpha_\text{KL}(\mathcal{L}_\text{KL}^r + \mathcal{L}_\text{KL}^g) + \alpha_\text{app}(\mathcal{L}_\text{app}^r + \mathcal{L}_\text{app}^g)+\alpha_\text{ad}(\mathcal{L}_\text{ad}^r + \mathcal{L}_\text{ad}^g)
\label{eq:total_loss}
\end{equation}
where the $\mathcal{L}_\text{KL}$ group regularizes consistency between pairs of distributions in terms of KL divergences, the $\mathcal{L}_\text{app}$ group encourages appearance matching fidelity, and while the $\mathcal{L}_\text{ad}$ group forces sampled images to fit in with the training set distribution. Each of the groups has a separate term for the reconstructive and generative paths. 

\paragraph{Distributive Regularization}
The typical interpretation of the KL divergence term in a VAE is that it regularizes the learned importance sampling function $q_\psi(\cdot|\mathbf{I}_c)$ to a fixed latent prior $p(\mathbf{z}_c)$. Defining as Gaussians, we get:
\begin{equation}
\mathcal{L}_\text{KL}^{r,(i)} = -\text{KL}(q_\psi(\mathbf{z}|I_c^{(i)})||\mathcal{N}_m(\mathbf{0},\sigma^2(n)\mathbf{I}))
\end{equation} 

For the generative path, the appropriate interpretation is \emph{reversed}: the learned conditional prior  $p_\phi(\cdot|\mathbf{I}_m)$, also a Gaussian, is regularized to $q_\psi(\cdot|\mathbf{I}_c)$.
\begin{equation}
\mathcal{L}_\text{KL}^{g,(i)} = -\text{KL}(q_\psi(\mathbf{z}|I_c^{(i)}))||p_\phi(\mathbf{z}|I_m^{(i)})))
\end{equation}
Note that the conditional prior only uses $\mathbf{I}_m$, while the importance function has access to the hidden $\mathbf{I}_c$.

\paragraph{Appearance Matching Loss}
The likelihood term $p^r_\theta(\mathbf{I}_c|\mathbf{z}_c,\mathbf{I}_m)$ may be interpreted as probabilistically encouraging appearance matching to the hidden $\mathbf{I}_c$. However, our framework also auto-encodes the visible $\mathbf{I}_m$ deterministically, and the loss function needs to cater for this reconstruction. As such, the per-instance loss here is:
\begin{equation}
\mathcal{L}_\text{app}^{r, (i)} = ||I_\text{rec}^{(i)}-I_g^{(i)}||_1
\label{eq:app_loss_rec}
\end{equation} 
where $I_\text{rec}^{(i)}$=$G(z_c,f_m)$ and $I_g^{(i)}$ are the reconstructed and original full images respectively.
In contrast, for the generative path we ignore instance-specific appearance matching for $\mathbf{I}_c$, and only focus on reconstructing $\mathbf{I}_m$ (via $\mathbf{f}_m$):
\begin{equation}
\mathcal{L}_\text{app}^{g, (i)} = ||M*(I_\text{gen}^{(i)}-I_g^{(i)})||_1
\label{eq:app_loss_gen}
\end{equation}
where $I_\text{gen}^{(i)}$=$G(\tilde{z}_c,f_m)$ is the generated image from the $\tilde{z}_c$ sample, and $M$ is the binary mask selecting visible pixels. 

\paragraph{Adversarial Loss}
The formulation of $p^r_\theta(\mathbf{I}_c|\mathbf{z}_c,\mathbf{I}_m)$ and the instance-blind $p^g_\theta(\mathbf{I}_c|\mathbf{\tilde z}_c,\mathbf{I}_m)$ also incorporates the use of adversarially learned discriminators $D_1$ and $D_2$ to judge whether the generated images fit into the training set distribution. Inspired by \cite{bao2017cvae}, we use a mean feature match loss in the reconstructive path for the generator,
\begin{equation}
\mathcal{L}_\text{ad}^{r, (i)} = ||f_{D_{1}}(I_\text{rec}^{(i)}) - f_{D_{1}}(I_g^{(i)})||_2
\label{eq:ad_loss_rec}
\end{equation}
where $f_{D_{1}}(\cdot)$ is the feature output of the final layer of $D_1$. This encourages the original and reconstructed features in the discriminator to be close together. Conversely, the adversarial loss in the generative path for the generator is:
\begin{equation}
\mathcal{L}_\text{ad}^{g, (i)} = [D_2(I_\text{gen}^{(i)})-1]^2
\label{eq:ad_loss_gen}
\end{equation}
This is based on the generator loss in LSGAN \cite{mao2017least}, which performs better than the original GAN loss \cite{goodfellow2014generative} in our scenario. The discriminator loss for both $D_1$ and $D_2$ is also based on LSGAN.

\begin{figure}[tb!]
	\centering
	\includegraphics[width=\linewidth]{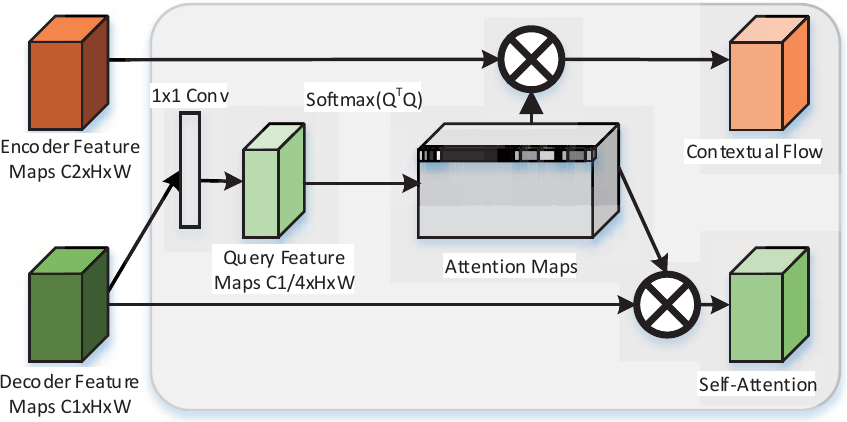}
	\caption{Our short+long term attention layer. The attention map is directly computed on the decoder features. After obtaining the self-attention scores, we use these to compute self-attention on decoder features, as well as contextual flow on encoder features. }
	\label{fig:attention}	
\end{figure}

\subsection{Short+Long Term Attention}

%A weakness of purely convolutional operations is that they have limited spatial ranges, and cannot exploit long-range correlation. Attention maps help solve this by copying information from visible patches to hidden patches \cite{yu2018generative,Yan_2018_ECCV,song2018contextual}, but the outcome depends on finding relevant features between visible and hidden regions in those work.

Extending beyond the Self-Attention GAN \cite{zhang2018self}, we propose not only to use the self-attention map within a decoder layer to harness \emph{distant spatial context}, but also to further capture \emph{feature-feature context} between encoder and decoder layers.
Our \emph{key novel insight} is: doing so would allow the network a choice of attending to the finer-grained features in the encoder or the more semantically generative features in the decoder, depending on circumstances.

Our proposed structure is shown in fig.~\ref{fig:attention}. We first calculate the self-attention map from the features $\mathbf{f}_d$ of a decoder middle layer, using the attention score of:
\begin{equation}
\beta_{j,i} = \frac{\exp(s_{ij})}{\sum_{i=1}^{N}\exp(s_{ij})},\mbox{where } s_{ij}=Q(f_{di})^TQ(f_{dj}),
\end{equation}
$N$ is the number of pixels, $Q(\mathbf{f}_d)$=$\mathbf{W}_q\mathbf{f}_d$, and $\mathbf{W}_q$ is a 1x1 convolution filter.
%to learn the \emph{query} features.
This leads to the short-term intra-layer attention feature (\textbf{self-attention} in fig.~\ref{fig:attention}) and the output $\mathbf{y}_d$:
\begin{equation}
c_{dj} = \sum_{i=1}^{N}\beta_{j,i}f_{di}
\;, \hspace{0.5cm}
\mathbf{y}_d= \gamma_d\mathbf{c}_d+\mathbf{f}_d
\end{equation}
where, following \cite{zhang2018self}, we use a scale parameter $\gamma_d$ to balance the weights between $\mathbf{c}_d$ and $\mathbf{f}_d$. The initial value of $\gamma_d$ is set to zero. 
In addition, for attending to features $\mathbf{f}_e$ from an encoder layer, we have a long-term inter-layer attention feature (\textbf{contextual flow} in fig.~\ref{fig:attention}) and the output $\mathbf{y}_e$:
\begin{equation}
c_{ej}=\sum_{i=1}^{N}\beta_{j,i}f_{ei}
\;, \hspace{0.5cm}
\mathbf{y}_e= \gamma_e(1-\mathbf{M})\mathbf{c}_e+\mathbf{M}\mathbf{f}_e
\end{equation}
As before, a scale parameter $\gamma_e$ is used to combine the encoder feature $\mathbf{f}_e$ and the attention feature $\mathbf{c}_e$. However, unlike the decoder feature $\mathbf{f}_d$ which has information for generating a full image, the encoder feature $\mathbf{f}_e$ only represents visible parts $\mathbf{I}_m$. Hence, a binary mask $\mathbf{M}$ (holes=0) is used. Finally, both the short and long term attention features are aggregated and fed into further decoder layers. 

\section{Experimental Results}
We evaluated our proposed model on four datasets including Paris \cite{doersch2012makes}, CelebA-HQ \cite{liu2015deep,karras2017progressive}, Places2 \cite{zhou2018places}, and ImageNet \cite{russakovsky2015imagenet} using the original training and test splits for those datasets. Since our model can generate multiple outputs, we sampled $50$ images for each masked image, and chose the top 10 results based on the discriminator scores. We trained our models for both regular and irregular holes. For brevity, we refer to our method as \textbf{PICNet}. We provide \href{https://github.com/lyndonzheng/Pluralistic}{PyTorch} implementations and \href{http://www.chuanxiaz.com/project/pluralistic/}{interactive demo}.

\subsection{Implementation Details}

Our generator and discriminator networks are inspired by SA-GAN \cite{zhang2018self}, but with several important modifications, including the short+long term attention layer. Furthermore, inspired by the growing-GAN \cite{karras2017progressive}, multi-scale output is applied to make the training faster. 

The image completion network, implemented in Pytorch v0.4.0, contains 6M trainable parameters. During optimization, the weights of different losses are set to $\alpha_\text{KL}=\alpha_\text{rec}$=20, $\alpha_\text{ad}$=1. We used Orthogonal Initialization \cite{saxe2013exact} and the Adam solver \cite{kingma2014adam}. All networks were trained from scratch, with a fixed learning rate of $\lambda$=$10^\text{-4}$. Details are in the supplemental section~\ref{experiment}.

\subsection{Comparison with Existing Work} 

\paragraph{Quantitative Comparisons}
Quantitative evaluation is hard for the pluralistic image completion task, as our goal is to get diverse but reasonable solutions for one masked image. The original image is only one solution of many, and comparisons should not be made based on just this image. 

However, just for the sake of obtaining quantitative measures, we will assume that one of our top 10 samples (ranked by the discriminator) will be close to the original ground truth, and select the single sample with the best balance of quantitative measures for comparison. The comparison is conducted on ImageNet $20,000$ test images, with quantitative measures of mean $\ell_1$ loss, peak signal-to-noise ration (PSNR), total variation (TV), and Inception Score (IS) \cite{salimans2016improved}. We used a $128\times128$ mask in the center.

\begin{table}[h]
	\begin{center}
		\begin{tabular}{|c|c|c|c|c|}
			\hline
			Method & $\ell_1$ loss & PSNR & TV loss & IS \\
			\hline
			GL \cite{iizuka2017globally}& 15.32 & 19.36 & 13.97 & 24.31\\
			CA \cite{yu2018generative}& 13.57 & 19.22 & 19.55 & {\bf 28.80}\\
			PICNet-regular& {\bf 12.91} & {\bf 20.10} & {\bf 12.18} & 24.90\\
			\hline
		\end{tabular}
	\end{center}
	\caption{Quantitative comparison with state-of-the-art. For center masks, our model was trained on regular holes.}
	\label{quantitative_comparisons}
\end{table}

\begin{figure*}[tb!]
	\centering
	\includegraphics[width=\textwidth]{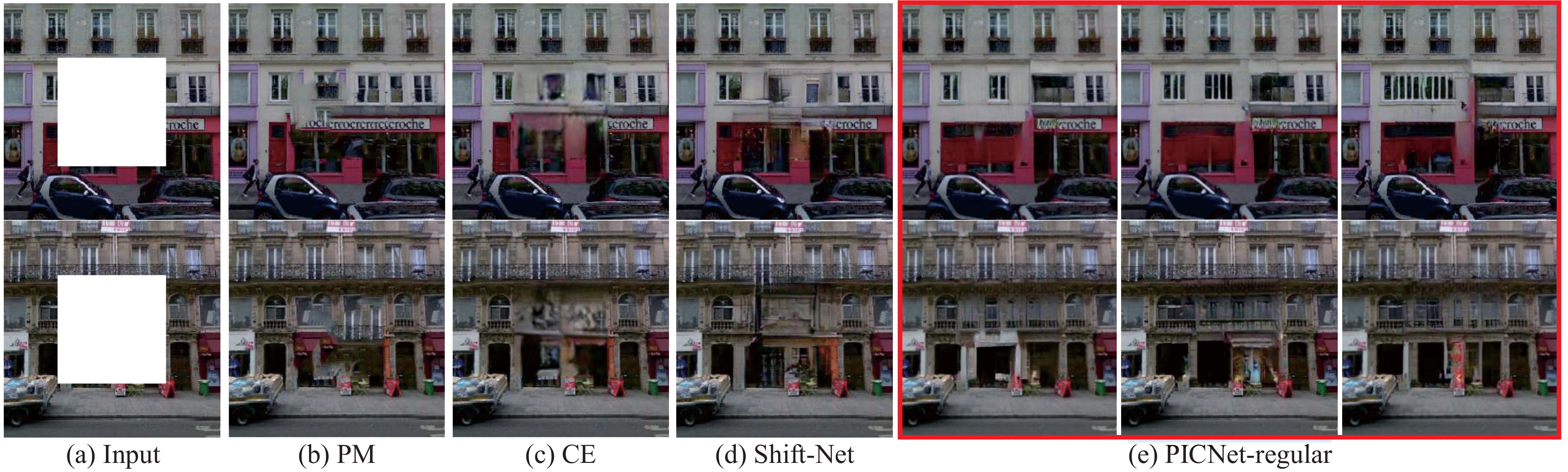}
	\caption{Comparison of our model with PatchMatch(PM) \cite{barnes2009patchmatch}, Context Encoder(CE)  \cite{pathak2016context} and Shift-Net \cite{Yan_2018_ECCV} on images taken from the Paris  \cite{doersch2012makes} test set for center region completion. Best viewed by zooming in.}
	\label{fig:result_building}	
\end{figure*}

\begin{figure*}[tb!]
	\centering
	\includegraphics[width=\textwidth]{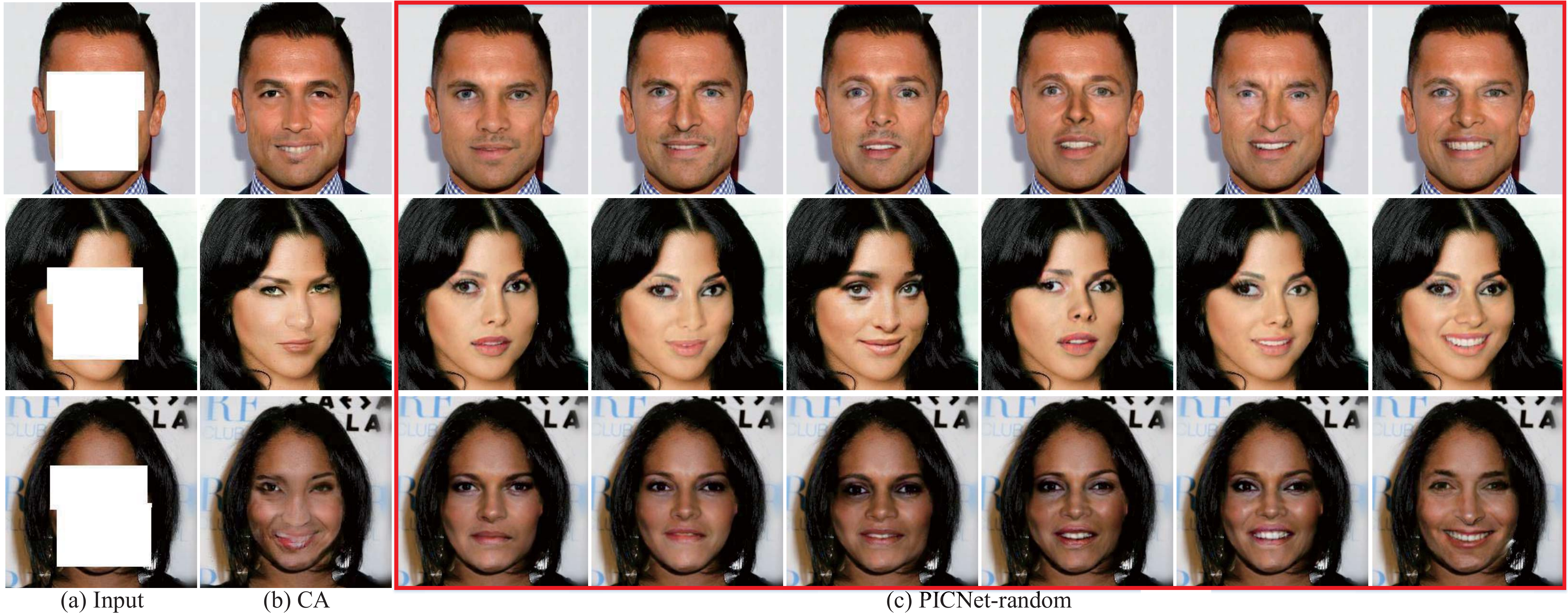}
	\caption{Comparison of our model with Contextual Attention(CA) \cite{yu2018generative} on CelebA-HQ. Best viewed by zooming in. }
	\label{fig:result_face}
\end{figure*}

\begin{figure*}[tb!]
	\centering
	\includegraphics[width=\textwidth]{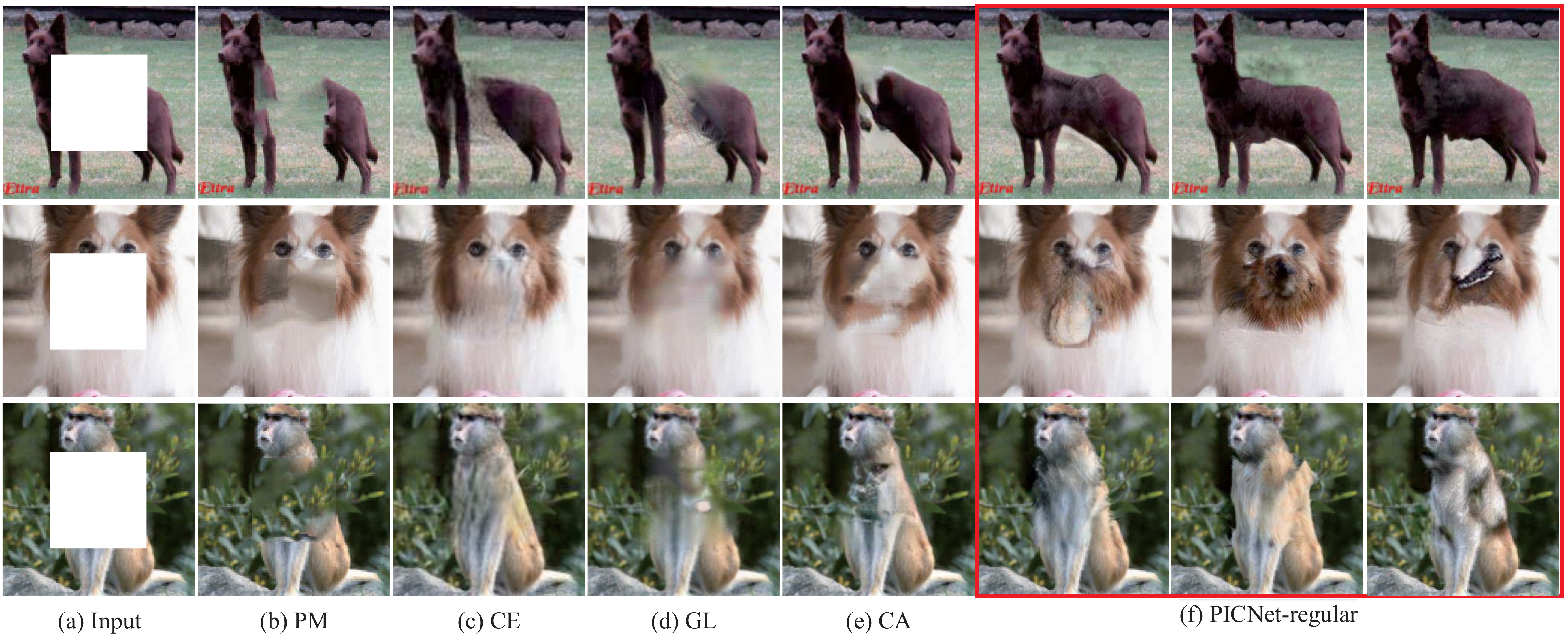}
	\caption{Qualitative results and comparisons with the PM, CE, Global and Local(GL) \cite{iizuka2017globally} and CA on the ImageNet validation set.}
	\label{fig:result_scene}	
\end{figure*}

\noindent\textbf{Qualitative Comparisons\quad}First, we show the results in fig.~\ref{fig:result_building} on the Paris dataset \cite{doersch2012makes}. For fair comparison among learning-based methods, we only compared with those trained on this dataset. PatchMatch \cite{barnes2009patchmatch} worked by copying similar patches from visible regions and obtained good results on this dataset with repetitive structures. Context Encoder (CE) \cite{pathak2016context} generated reasonable structures with blurry textures. Shift-Net \cite{Yan_2018_ECCV} made improvements by feature copying. Compared to these, our model not only generated more natural images, but also with multiple solutions, \eg different numbers of windows and varying door sizes.

Next, we evaluated our methods on CelebA-HQ face dataset, with fig.~\ref{fig:result_face} showing examples with large regular holes to highlight the diversity of our output.
Context Attention (CA) \cite{yu2018generative} generated reasonable completion for many cases, but for each masked input they were only able to generate a single result; furthermore, on some occasions, the single solution may be poor. Our model produced various plausible results by sampling from the latent space conditional prior. 

%Note that during the training, no any other information is used in our model. 

Finally, we report the performance on the more challenging ImageNet dataset by comparing to the previous PatchMatch \cite{barnes2009patchmatch}, CE \cite{pathak2016context}, GL \cite{iizuka2017globally} and CA \cite{yu2018generative}. Different from the CE and GL models that were trained on the $100$k subset of training images of ImageNet, our model is directly trained on original ImageNet training dataset with all images resized to $256\times256$. Visual results on a variety of objects from the validation set are shown in fig.~\ref{fig:result_scene}. Our model was able to infer the content quite effectively.  

%These visual test images were those chosen in \cite{iizuka2017globally}. The CE results were obtained from the authors'  website, while we used publicly released trained models for GL and CA. As shown in fig.~\ref{fig:result_scene},

\begin{figure}[tb!]
	\setlength{\belowcaptionskip}{-10pt} 
	\centering
	\includegraphics[width=\linewidth]{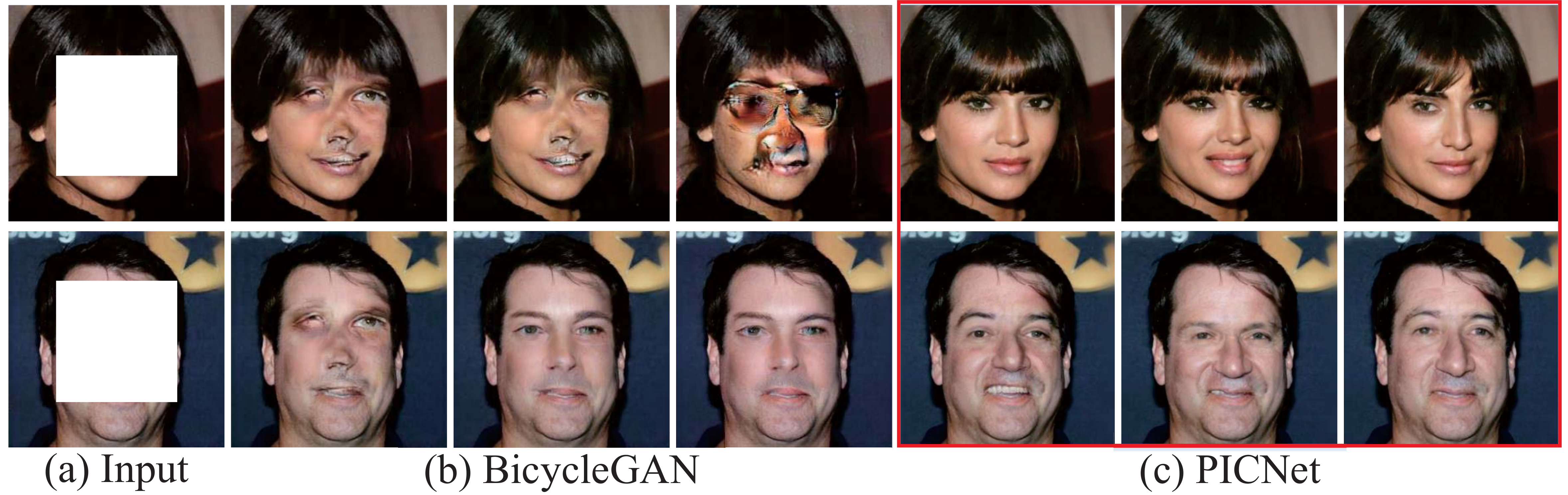}
	\caption{Comparison of our Pluralistic model with BicycleGAN.}
	\label{fig:bicycleGAN}	
\end{figure} 

\begin{figure}[tb!]
	\centering
	\includegraphics[width=\linewidth]{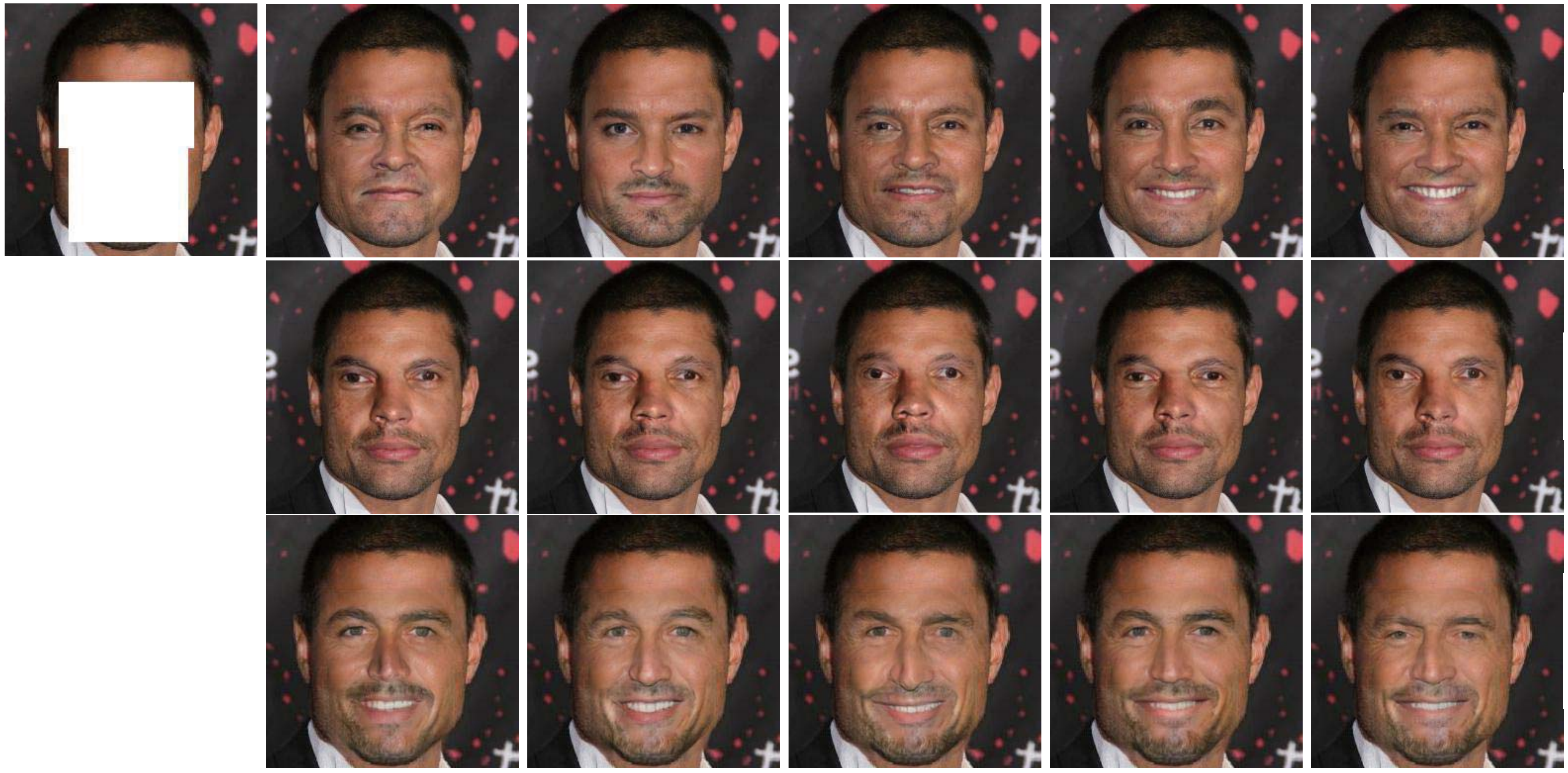}
	\caption{Comparison of training with different strategies: ours (top), CVAE (middle), instance-blind (bottom).}
	\label{fig:ablation_paths}	
\end{figure}

\subsection{Ablation Study}

%\paragraph{Regions-Related Distribution {\emph vs.} Fixed Distribution}

\paragraph{Our PICNet vs CVAE vs ``Instance Blind'' vs BicycleGAN}
We investigated the influence of using our two-path training structure in comparison to other variants such as the CVAE \cite{sohn2015learning} and ``instance blind'' structures in fig.~\ref{fig:coarse_framework}. We trained the three models using common parameters. As shown in fig.~\ref{fig:ablation_paths}, for the CVAE, even after sampling from the latent prior distribution, the outputs were almost identical, as the conditional prior learned is narrowly centered at the maximum latent likelihood solution. As for ``instance blind'', if reconstruction loss was used only on visible pixels, the training may become unstable. If we used reconstruction loss on the full generated image,  there is also little variation as the framework has likely learned to ignore the sampling and predicted a deterministic outcome purely from $\mathbf{I}_m$.  

We also trained and tested BicycleGAN \cite{zhu2017toward} for center masks. As is obvious in fig.~\ref{fig:bicycleGAN}, BicycleGAN is not directly suitable, leading to poor results or minimal variation.

\noindent\textbf{Diversity Measure\quad} We computed diversity scores using the LPIPS metric reported in [46]. The average score is calculated between 50K pairs generated from a sampling of 1K center-masked images. $\mathbf{I}_{out}$ and $\mathbf{I}_{out(m)}$ are the full output and mask-region output, respectively. While [46] obtained relatively higher diversity scores (still lower than ours), most of their generated images look unnatural (fig.~\ref{fig:bicycleGAN}).

\begin{table}[tb!]
	\begin{center}
		\begin{tabular}{|c|c|c|}
			\hline
			\multicolumn{1}{|c|}{} &  \multicolumn{2}{|c|}{Diversity (LPIPS)} \\
			\hline
			Method & $\mathbf{I}_{out}$ & $\mathbf{I}_{out(m)}$ \\
			%\hline
			%Random real images & & \\
			\hline
			CVAE & 0.004& 0.014\\
			\hline
			Instance Blind & 0.015& 0.049\\
			\hline
			BicycleGAN [46] & 0.027& 0.060\\
			\hline
			PICNet-Pluralistic & {\bf0.029} & {\bf0.088}\\
			\hline
		\end{tabular}
	\end{center}
	\caption{Quantitative comparisons of diversity.}
	\label{diversity_comparisons}
\end{table}

\begin{figure}[tb!]
	\centering
	\includegraphics[width=\linewidth]{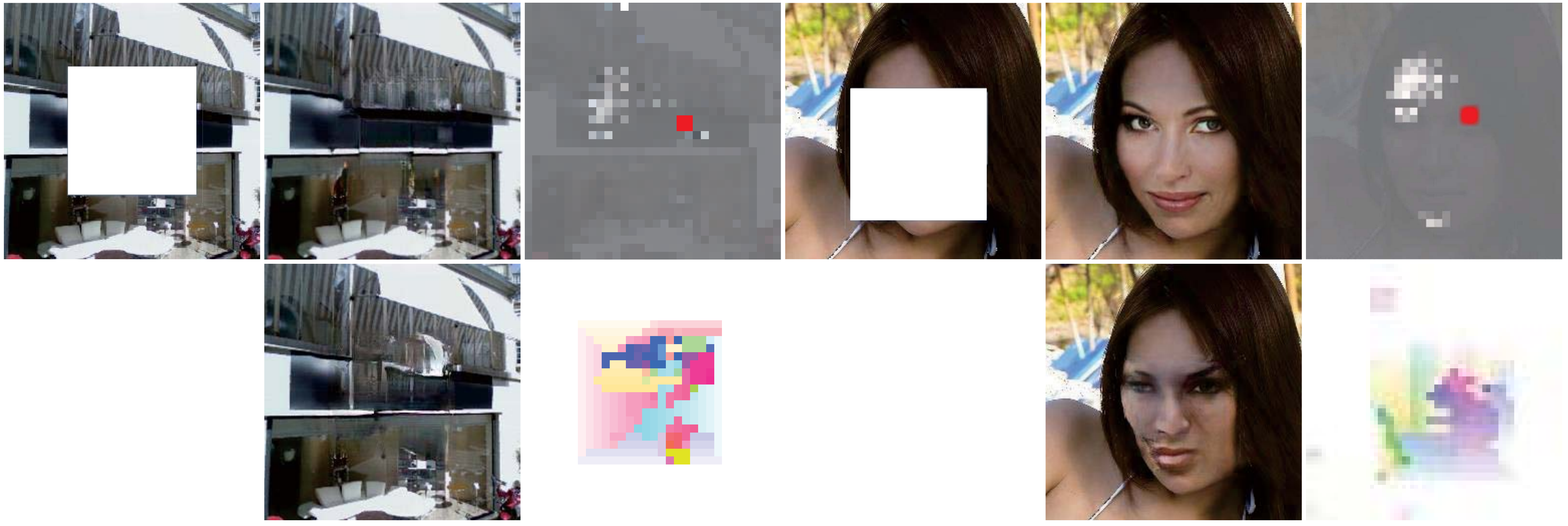}
	\caption{Visualization of attention map using different attention modules: ours (top), contextual attention (bottom). We highlight the most-attended regions for the query position (red point).}
	\label{fig:ablation_attention}	
\end{figure}

\noindent\textbf{Short+Long Term Attention vs Contextual Attention \quad}
We visualized our attention maps as in \cite{zhang2018self}. To compare to the contextual attention (CA) layer \cite{yu2018generative}, we retrained CA on the Paris dataset via the authors' code, and used their publicly released face model. The CA attention maps are presented in their color-directional format. As shown in fig.~\ref{fig:ablation_attention}, our short+long term attention layer borrowed features from different positions with varying attention weights, rather than directly copying similar features from just one visible position. For the building scene, CA's results were of similar high quality to ours, due to the repeated structures present. However for a face with a large mask, CA was unable to borrow features for the hidden content (\eg mouth, eyes) from visible regions, with poor output. Our attention map is able to utilize both decoder features (which do not have masked parts) and encoder features as appropriate.

%\subsection{limitations}

\section{Conclusion}

We proposed a novel dual pipeline training architecture for pluralistic image completion. Unlike existing methods, our framework can generate multiple diverse solutions with plausible content for a single masked input. The experimental results demonstrate this prior-conditional lower bound coupling is significant for conditional image generation.  We also introduced an enhanced short+long term attention layer which improves realism. Experiments on a variety of datasets showed that our multiple solutions were diverse and of high-quality, especially for large holes.

\paragraph{\bf Acknowledgements} This research is supported by the BeingTogether Centre, a collaboration between Nanyang Technological University (NTU) Singapore and University of North Carolina (UNC) at Chapel Hill. The BeingTogether Centre is supported by the National Research Foundation, Prime Minister’s Office, Singapore under its International Research Centres in Singapore Funding Initiative. This research was also conducted in collaboration with Singapore Telecommunications Limited and partially supported by the Singapore Government through the Industry Alignment Fund ‐- Industry Collaboration Projects Grant.

{\small
\bibliographystyle{ieee}
\bibliography{egbib}

\begin{thebibliography}{10}\itemsep=-1pt

\bibitem{ballester2001filling}
Coloma Ballester, Marcelo Bertalmio, Vicent Caselles, Guillermo Sapiro, and
  Joan Verdera.
\newblock Filling-in by joint interpolation of vector fields and gray levels.
\newblock {\em IEEE transactions on image processing}, 10(8):1200--1211, 2001.

\bibitem{bao2017cvae}
Jianmin Bao, Dong Chen, Fang Wen, Houqiang Li, and Gang Hua.
\newblock Cvae-gan: Fine-grained image generation through asymmetric training.
\newblock In {\em 2017 IEEE International Conference on Computer Vision
  (ICCV)}, pages 2764--2773. IEEE, 2017.

\bibitem{barnes2009patchmatch}
Connelly Barnes, Eli Shechtman, Adam Finkelstein, and Dan~B Goldman.
\newblock Patchmatch: A randomized correspondence algorithm for structural
  image editing.
\newblock {\em ACM Transactions on Graphics (ToG)}, 28:24, 2009.

\bibitem{bertalmio2000image}
Marcelo Bertalmio, Guillermo Sapiro, Vincent Caselles, and Coloma Ballester.
\newblock Image inpainting.
\newblock In {\em Proceedings of the 27th annual conference on Computer
  graphics and interactive techniques}, pages 417--424. ACM
  Press/Addison-Wesley Publishing Co., 2000.

\bibitem{bertalmio2003simultaneous}
Marcelo Bertalmio, Luminita Vese, Guillermo Sapiro, and Stanley Osher.
\newblock Simultaneous structure and texture image inpainting.
\newblock {\em IEEE transactions on image processing}, 12(8):882--889, 2003.

\bibitem{chen2018high}
Zeyuan Chen, Shaoliang Nie, Tianfu Wu, and Christopher~G Healey.
\newblock High resolution face completion with multiple controllable attributes
  via fully end-to-end progressive generative adversarial networks.
\newblock {\em arXiv preprint arXiv:1801.07632}, 2018.

\bibitem{criminisi2003object}
Antonio Criminisi, Patrick Perez, and Kentaro Toyama.
\newblock Object removal by exemplar-based inpainting.
\newblock In {\em Computer Vision and Pattern Recognition, 2003. Proceedings.
  2003 IEEE Computer Society Conference on}, volume~2, pages II--II. IEEE,
  2003.

\bibitem{criminisi2004region}
Antonio Criminisi, Patrick P{\'e}rez, and Kentaro Toyama.
\newblock Region filling and object removal by exemplar-based image inpainting.
\newblock {\em IEEE Transactions on image processing}, 13(9):1200--1212, 2004.

\bibitem{doersch2012makes}
Carl Doersch, Saurabh Singh, Abhinav Gupta, Josef Sivic, and Alexei Efros.
\newblock What makes paris look like paris?
\newblock {\em ACM Transactions on Graphics}, 31(4), 2012.

\bibitem{Eslami2018}
S.~M.~Ali Eslami, Danilo Jimenez~Rezende, Frederic Besse, Fabio Viola, Ari~S.
  Morcos, Marta Garnelo, Avraham Ruderman, Andrei~A. Rusu, Ivo Danihelka, Karol
  Gregor, David~P. Reichert, Lars Buesing, Theophane Weber, Oriol Vinyals, Dan
  Rosenbaum, Neil Rabinowitz, Helen King, Chloe Hillier, Matt Botvinick, Daan
  Wierstra, Koray Kavukcuoglu, and Demis Hassabis.
\newblock Neural scene representation and rendering.
\newblock {\em Science}, 360(6394):1204--1210, 2018.

\bibitem{goodfellow2014generative}
Ian Goodfellow, Jean Pouget-Abadie, Mehdi Mirza, Bing Xu, David Warde-Farley,
  Sherjil Ozair, Aaron Courville, and Yoshua Bengio.
\newblock Generative adversarial nets.
\newblock In {\em Advances in neural information processing systems}, pages
  2672--2680, 2014.

\bibitem{hays2007scene}
James Hays and Alexei~A Efros.
\newblock Scene completion using millions of photographs.
\newblock In {\em ACM Transactions on Graphics (TOG)}, volume~26, page~4. ACM,
  2007.

\bibitem{huang2014image}
Jia-Bin Huang, Sing~Bing Kang, Narendra Ahuja, and Johannes Kopf.
\newblock Image completion using planar structure guidance.
\newblock {\em ACM Transactions on graphics (TOG)}, 33(4):129, 2014.

\bibitem{iizuka2017globally}
Satoshi Iizuka, Edgar Simo-Serra, and Hiroshi Ishikawa.
\newblock Globally and locally consistent image completion.
\newblock {\em ACM Transactions on Graphics (TOG)}, 36(4):107, 2017.

\bibitem{isola2017image}
Phillip Isola, Jun-Yan Zhu, Tinghui Zhou, and Alexei~A Efros.
\newblock Image-to-image translation with conditional adversarial networks.
\newblock In {\em 2017 IEEE Conference on Computer Vision and Pattern
  Recognition (CVPR)}, pages 5967--5976. IEEE, 2017.

\bibitem{jaderberg2015spatial}
Max Jaderberg, Karen Simonyan, Andrew Zisserman, et~al.
\newblock Spatial transformer networks.
\newblock In {\em Advances in neural information processing systems}, pages
  2017--2025, 2015.

\bibitem{karras2017progressive}
Tero Karras, Timo Aila, Samuli Laine, and Jaakko Lehtinen.
\newblock Progressive growing of gans for improved quality, stability, and
  variation.
\newblock {\em arXiv preprint arXiv:1710.10196}, 2017.

\bibitem{kingma2014adam}
Diederik~P Kingma and Jimmy Ba.
\newblock Adam: A method for stochastic optimization.
\newblock {\em arXiv preprint arXiv:1412.6980}, 2014.

\bibitem{kingma2013auto}
Diederik~P Kingma and Max Welling.
\newblock Auto-encoding variational bayes.
\newblock {\em arXiv preprint arXiv:1312.6114}, 2013.

\bibitem{kohler2014mask}
Rolf K{\"o}hler, Christian Schuler, Bernhard Sch{\"o}lkopf, and Stefan
  Harmeling.
\newblock Mask-specific inpainting with deep neural networks.
\newblock In {\em German Conference on Pattern Recognition}, pages 523--534.
  Springer, 2014.

\bibitem{lee2018diverse}
Hsin-Ying Lee, Hung-Yu Tseng, Jia-Bin Huang, Maneesh Singh, and Ming-Hsuan
  Yang.
\newblock Diverse image-to-image translation via disentangled representations.
\newblock In {\em European Conference on Computer Vision (ECCV)}, 2018.

\bibitem{levin2003learning}
Anat Levin, Assaf Zomet, and Yair Weiss.
\newblock Learning how to inpaint from global image statistics.
\newblock In {\em null}, page 305. IEEE, 2003.

\bibitem{li2017generative}
Yijun Li, Sifei Liu, Jimei Yang, and Ming-Hsuan Yang.
\newblock Generative face completion.
\newblock In {\em Computer Vision and Pattern Recognition (CVPR), 2017 IEEE
  Conference on}, pages 5892--5900. IEEE, 2017.

\bibitem{Liu_2018_ECCV}
Guilin Liu, Fitsum~A. Reda, Kevin~J. Shih, Ting-Chun Wang, Andrew Tao, and
  Bryan Catanzaro.
\newblock Image inpainting for irregular holes using partial convolutions.
\newblock In {\em Proceedings of the European Conference on Computer Vision
  (ECCV)}, September 2018.

\bibitem{liu2015deep}
Ziwei Liu, Ping Luo, Xiaogang Wang, and Xiaoou Tang.
\newblock Deep learning face attributes in the wild.
\newblock In {\em Proceedings of the IEEE International Conference on Computer
  Vision}, pages 3730--3738, 2015.

\bibitem{mao2017least}
Xudong Mao, Qing Li, Haoran Xie, Raymond~YK Lau, Zhen Wang, and Stephen~Paul
  Smolley.
\newblock Least squares generative adversarial networks.
\newblock In {\em Computer Vision (ICCV), 2017 IEEE International Conference
  on}, pages 2813--2821. IEEE, 2017.

\bibitem{mathieu2015deep}
Michael Mathieu, Camille Couprie, and Yann LeCun.
\newblock Deep multi-scale video prediction beyond mean square error.
\newblock {\em arXiv preprint arXiv:1511.05440}, 2015.

\bibitem{park2017transformation}
Eunbyung Park, Jimei Yang, Ersin Yumer, Duygu Ceylan, and Alexander~C Berg.
\newblock Transformation-grounded image generation network for novel 3d view
  synthesis.
\newblock In {\em 2017 IEEE Conference on Computer Vision and Pattern
  Recognition (CVPR)}, pages 702--711. IEEE, 2017.

\bibitem{pathak2016context}
Deepak Pathak, Philipp Krahenbuhl, Jeff Donahue, Trevor Darrell, and Alexei~A
  Efros.
\newblock Context encoders: Feature learning by inpainting.
\newblock In {\em Proceedings of the IEEE Conference on Computer Vision and
  Pattern Recognition}, pages 2536--2544, 2016.

\bibitem{ren2015shepard}
Jimmy~SJ Ren, Li Xu, Qiong Yan, and Wenxiu Sun.
\newblock Shepard convolutional neural networks.
\newblock In {\em Advances in Neural Information Processing Systems}, pages
  901--909, 2015.

\bibitem{russakovsky2015imagenet}
Olga Russakovsky, Jia Deng, Hao Su, Jonathan Krause, Sanjeev Satheesh, Sean Ma,
  Zhiheng Huang, Andrej Karpathy, Aditya Khosla, Michael Bernstein, et~al.
\newblock Imagenet large scale visual recognition challenge.
\newblock {\em International Journal of Computer Vision}, 115(3):211--252,
  2015.

\bibitem{salimans2016improved}
Tim Salimans, Ian Goodfellow, Wojciech Zaremba, Vicki Cheung, Alec Radford, and
  Xi Chen.
\newblock Improved techniques for training gans.
\newblock In {\em Advances in Neural Information Processing Systems}, pages
  2234--2242, 2016.

\bibitem{saxe2013exact}
Andrew~M Saxe, James~L McClelland, and Surya Ganguli.
\newblock Exact solutions to the nonlinear dynamics of learning in deep linear
  neural networks.
\newblock {\em arXiv preprint arXiv:1312.6120}, 2013.

\bibitem{sohn2015learning}
Kihyuk Sohn, Honglak Lee, and Xinchen Yan.
\newblock Learning structured output representation using deep conditional
  generative models.
\newblock In {\em Advances in Neural Information Processing Systems}, pages
  3483--3491, 2015.

\bibitem{song2018contextual}
Yuhang Song, Chao Yang, Zhe Lin, Xiaofeng Liu, Qin Huang, Hao Li, and CC Jay.
\newblock Contextual-based image inpainting: Infer, match, and translate.
\newblock In {\em Proceedings of the European Conference on Computer Vision
  (ECCV)}, pages 3--19, 2018.

\bibitem{song2018spg}
Yuhang Song, Chao Yang, Yeji Shen, Peng Wang, Qin Huang, and C-C~Jay Kuo.
\newblock Spg-net: Segmentation prediction and guidance network for image
  inpainting.
\newblock {\em arXiv preprint arXiv:1805.03356}, 2018.

\bibitem{walker2016}
Jacob Walker, Carl Doersch, Abhinav Gupta, and Martial Hebert.
\newblock An uncertain future: Forecasting from static images using variational
  autoencoders.
\newblock In {\em European Conference on Computer Vision (ECCV)}, 2016.

\bibitem{Yan_2018_ECCV}
Zhaoyi Yan, Xiaoming Li, Mu Li, Wangmeng Zuo, and Shiguang Shan.
\newblock Shift-net: Image inpainting via deep feature rearrangement.
\newblock In {\em The European Conference on Computer Vision (ECCV)}, September
  2018.

\bibitem{yang2017high}
Chao Yang, Xin Lu, Zhe Lin, Eli Shechtman, Oliver Wang, and Hao Li.
\newblock High-resolution image inpainting using multi-scale neural patch
  synthesis.
\newblock In {\em The IEEE Conference on Computer Vision and Pattern
  Recognition (CVPR)}, volume~1, page~3, 2017.

\bibitem{yeh2017semantic}
Raymond~A Yeh, Chen Chen, Teck~Yian Lim, Alexander~G Schwing, Mark
  Hasegawa-Johnson, and Minh~N Do.
\newblock Semantic image inpainting with deep generative models.
\newblock In {\em Computer Vision and Pattern Recognition (CVPR), 2017 IEEE
  Conference on}, pages 6882--6890. IEEE, 2017.

\bibitem{yu2018free}
Jiahui Yu, Zhe Lin, Jimei Yang, Xiaohui Shen, Xin Lu, and Thomas~S Huang.
\newblock Free-form image inpainting with gated convolution.
\newblock {\em arXiv preprint arXiv:1806.03589}, 2018.

\bibitem{yu2018generative}
Jiahui Yu, Zhe Lin, Jimei Yang, Xiaohui Shen, Xin Lu, and Thomas~S Huang.
\newblock Generative image inpainting with contextual attention.
\newblock {\em arXiv preprint arXiv:1801.07892}, 2018.

\bibitem{zhang2018self}
Han Zhang, Ian Goodfellow, Dimitris Metaxas, and Augustus Odena.
\newblock Self-attention generative adversarial networks.
\newblock {\em arXiv preprint arXiv:1805.08318}, 2018.

\bibitem{zheng2018t2net}
Chuanxia Zheng, Tat-Jen Cham, and Jianfei Cai.
\newblock T2net: Synthetic-to-realistic translation for solving single-image
  depth estimation tasks.
\newblock In {\em Proceedings of the European Conference on Computer Vision
  (ECCV)}, pages 767--783, 2018.

\bibitem{zhou2018places}
Bolei Zhou, Agata Lapedriza, Aditya Khosla, Aude Oliva, and Antonio Torralba.
\newblock Places: A 10 million image database for scene recognition.
\newblock {\em IEEE transactions on pattern analysis and machine intelligence},
  40(6):1452--1464, 2018.

\bibitem{zhou2016view}
Tinghui Zhou, Shubham Tulsiani, Weilun Sun, Jitendra Malik, and Alexei~A Efros.
\newblock View synthesis by appearance flow.
\newblock In {\em European conference on computer vision}, pages 286--301.
  Springer, 2016.

\bibitem{zhu2017toward}
Jun-Yan Zhu, Richard Zhang, Deepak Pathak, Trevor Darrell, Alexei~A Efros,
  Oliver Wang, and Eli Shechtman.
\newblock Toward multimodal image-to-image translation.
\newblock In {\em Advances in Neural Information Processing Systems}, pages
  465--476, 2017.

\end{thebibliography}
}

\appendix\onecolumn
\renewcommand{\appendixname}{Appendix~\Alph{section}}
\renewcommand{\theequation}{\thesection.\arabic{equation}}
\setcounter{equation}{0}
\renewcommand{\thefigure}{\thesection.\arabic{figure}}
\setcounter{figure}{0}
\renewcommand{\thetable}{\thesection.\arabic{table}}
\setcounter{table}{0}
\newpage

\section{Additional Examples}\label{result}

We first show our results on center hole completion, in relation to those from other methods trained on corresponding datasets. As for random irregular and regular holes, we simply present our results so that readers may appreciate the multiple diverse results we can get with differently sized and shaped holes. Finally, we show the interesting application on face editing.

\subsection{Comparison with Existing Work on Center Hole Completion}

\begin{figure*}[h]
	\centering
	\includegraphics[width=\linewidth]{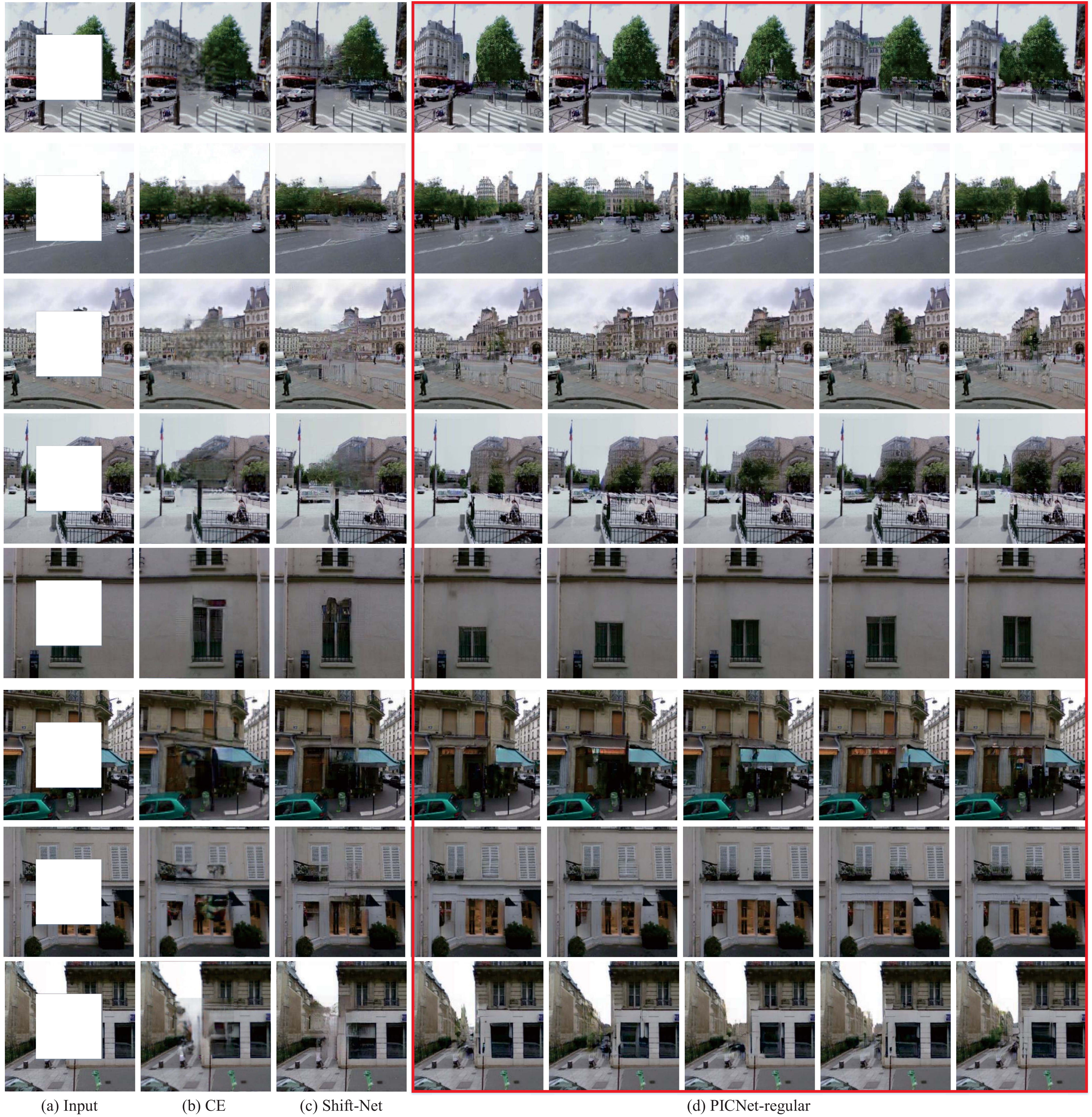}
	\caption{Additional results on the Paris variation set for center hole completion. This variation dataset contains 100 images, for which we obtained generally more realistic results than the existing methods of CE and Shift-Net. Furthermore, our multiple results had a diverse range of sizes, shapes, colors and textures. Best viewed by zooming in.}
	\label{fig:appendix_building_center}	
\end{figure*}

\begin{figure*}[h]
	\centering
	\includegraphics[width=\linewidth]{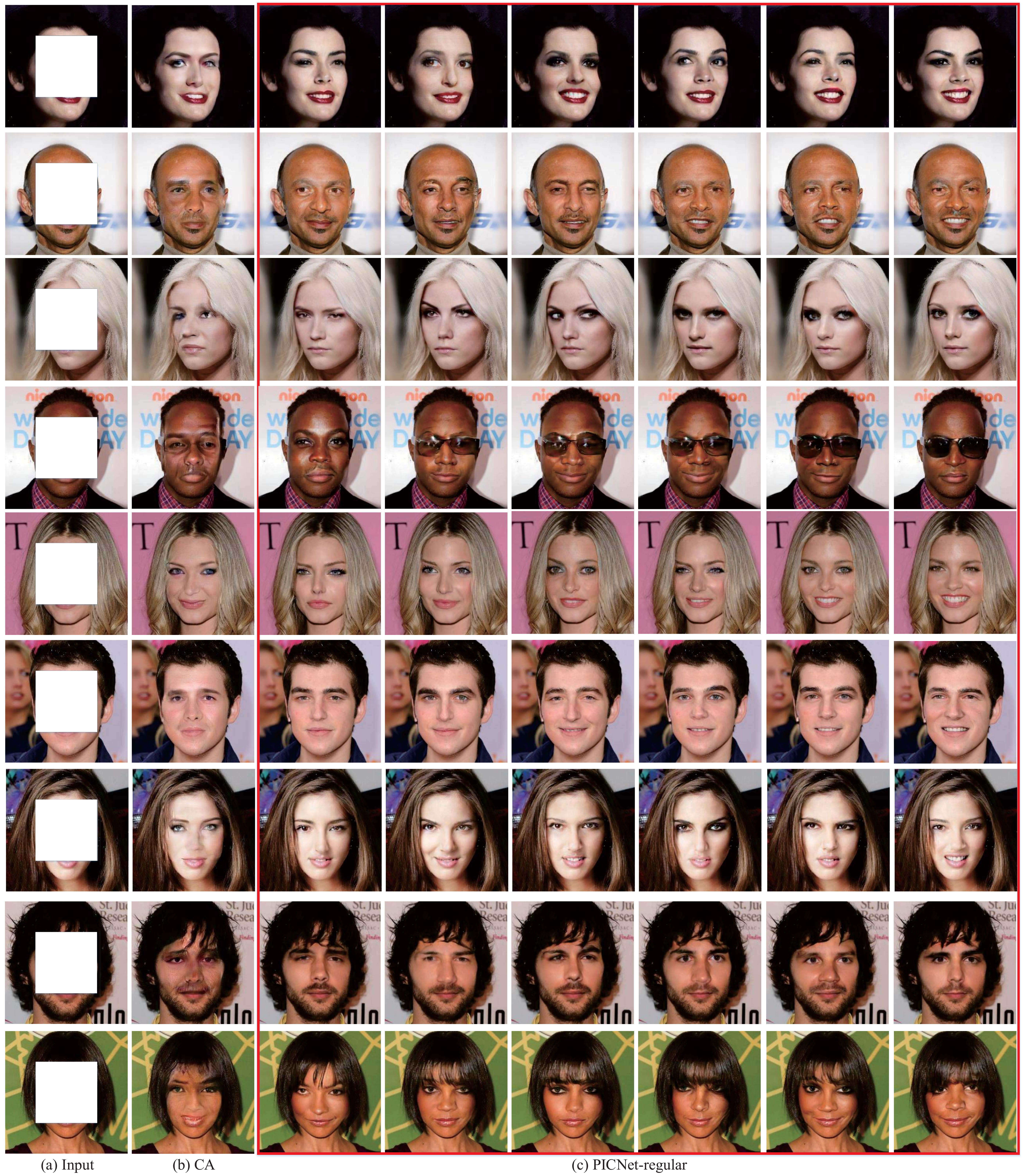}
	\caption{Additional results on the CelebA-HQ test set for center hole completion. Examples have different genders, skin tones, views and partial visible expressions. Since the occluded content in the large center holes was not repeated in visible regions, CA was unable to create results that were as visually realistic as ours. Moreover, our multiple outputs have different shapes, sizes and colors for eyes, noses and mouths. The details can be viewed by zooming in. Note that, no any other attribute labels (\eg smile) were applied in our approach.}
	\label{fig:appendix_face_center}	
\end{figure*}

\begin{figure*}[h]
	\centering
	\includegraphics[width=0.98\linewidth]{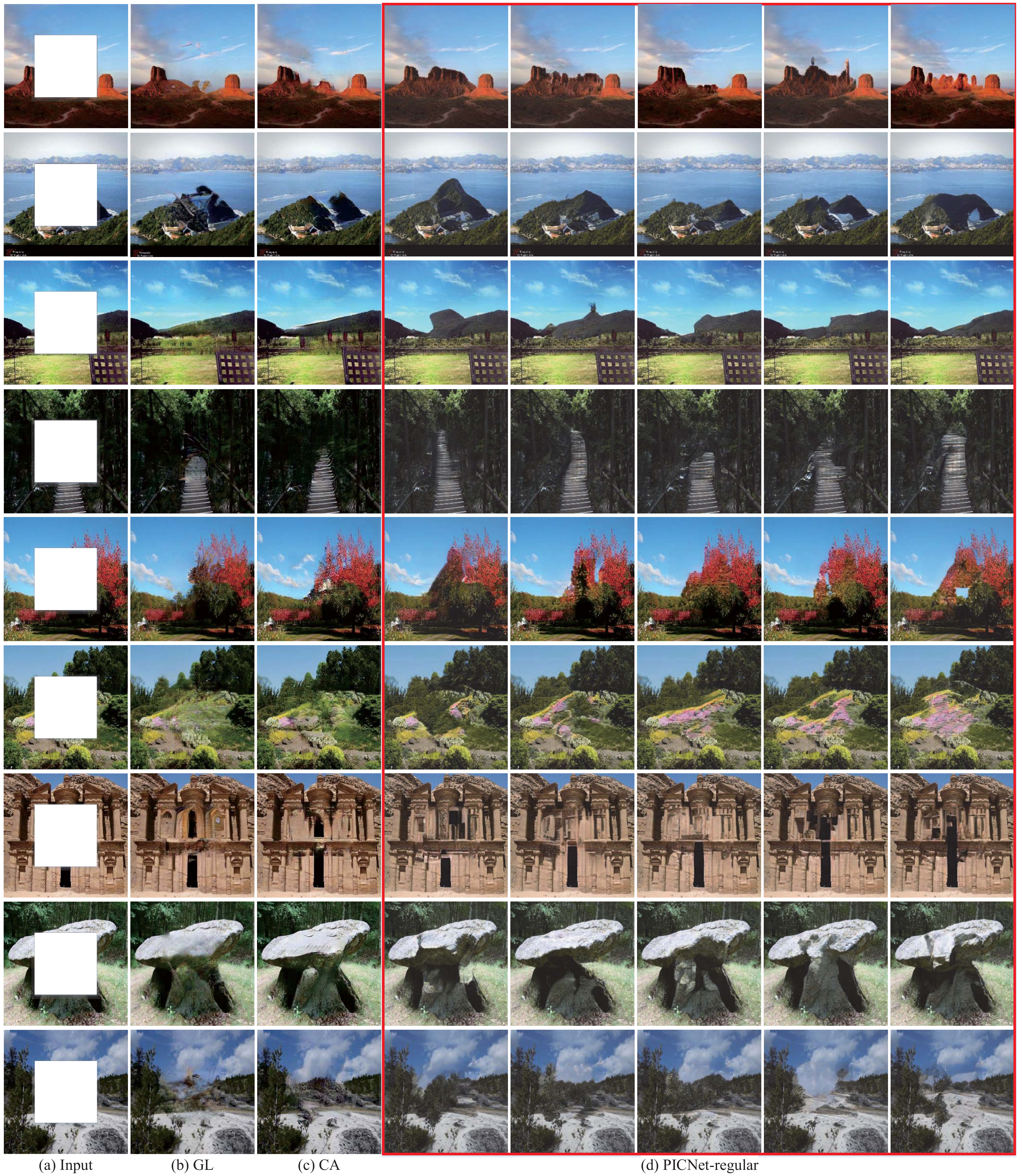}
	\caption{Additional results on the Places2 variation set for center hole completion. Compared with existing state-of-the-art methods, our model not only generated completion results of comparable quality, but also provided multiple plausible results, with different shapes, colors, textures and content. The shape variations in generating the various prominent hills are obvious. Some changes were at finer scale, \eg color changes of the flowers and texture changes in the boulder are better viewed by zooming in.}
	\label{fig:appendix_place_center}	
\end{figure*}

\begin{figure*}[h]
	\centering
	\includegraphics[width=0.98\linewidth]{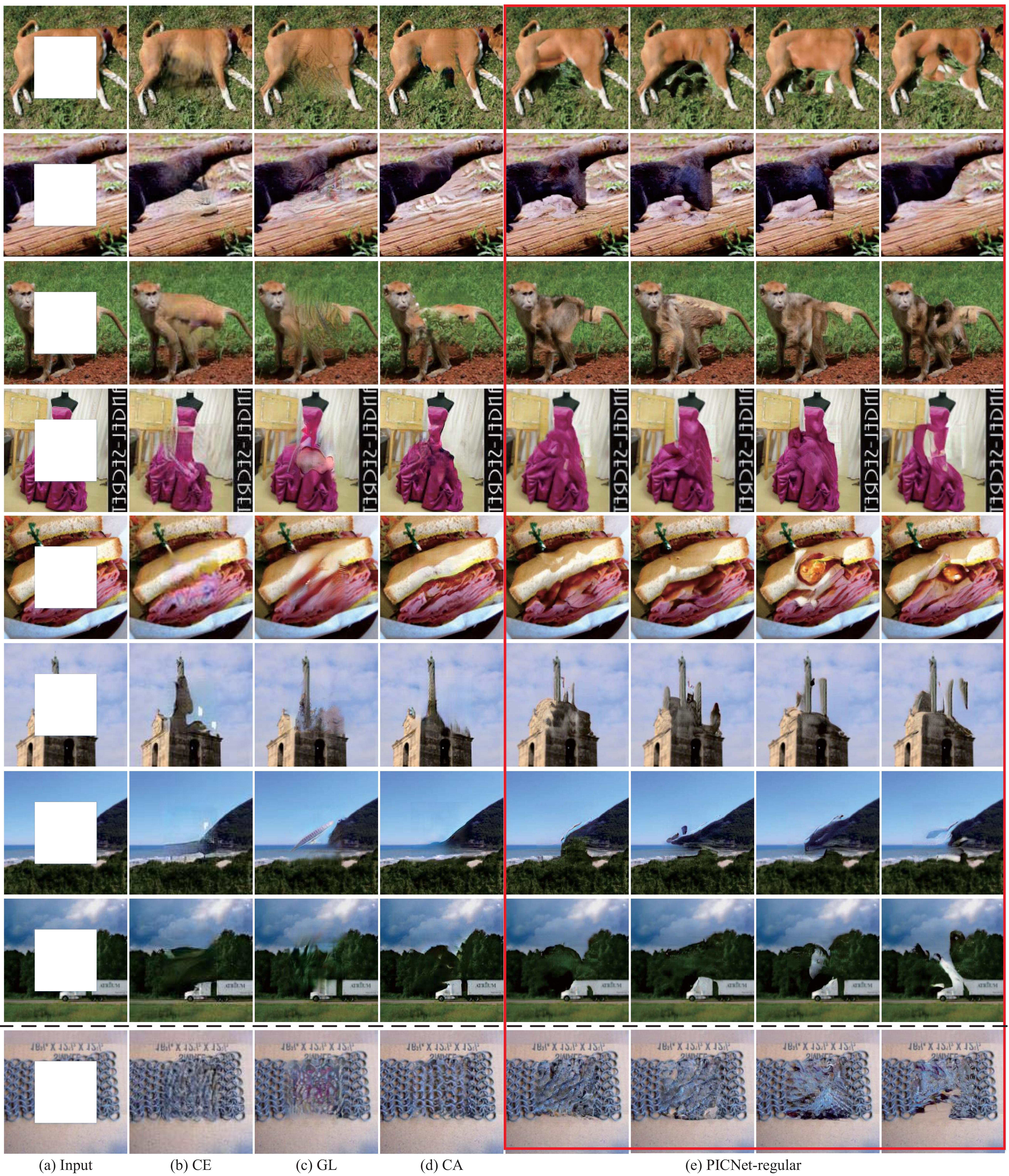}
	\caption{Additional results for center holes on the ImageNet variation set used in Context Encoder (CE). For our results, four completed images were selected and included failure cases in the last column. The first four rows show examples in which our model generated more visually realistic results than other methods. The next four rows show examples in which the methods all performed with similar realism, while the final row shows an example in which the Context Attention (CA) had the most realistic result. }
	\label{fig:appendix_imagenet_center}	
\end{figure*}

\subsection{Additional Results on Random and Irregular Hole Completion}

\begin{figure*}[h]
	\centering
	\includegraphics[width=\linewidth]{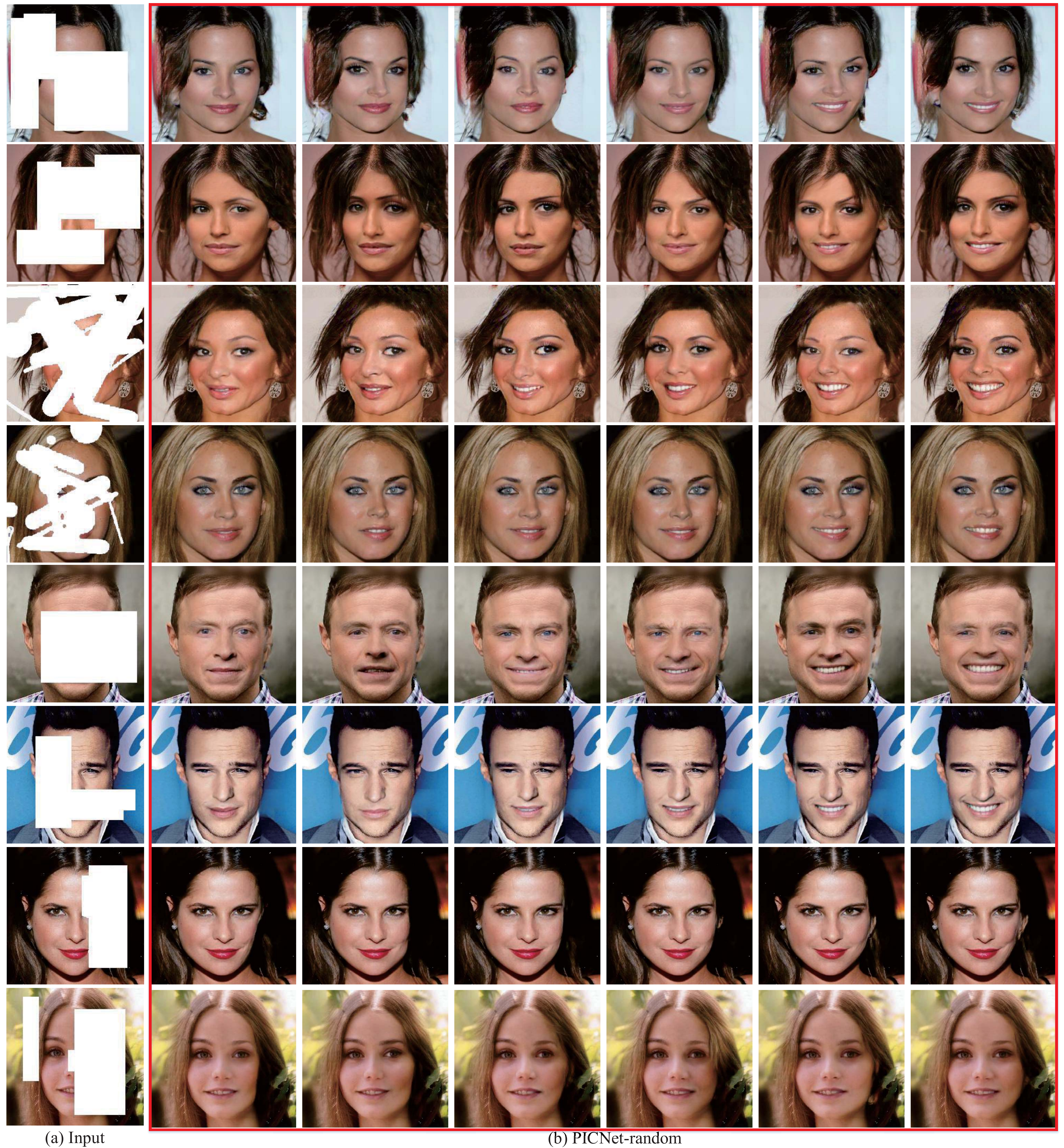}
	\caption{Additional results on the CelebA-HQ test set for random and irregular hole completion. One interesting observation is that natural facial symmetry exerts a strong constraint on the completion results. In the examples where both eyes and/or mouth are masked out, the completion results exhibit substantial variation for those facial features when sampled. However, when only one eye is masked out or half of the mouth is visible (last three rows), the completion results for the other eye or the other half of mouth have little variation when sampled. Even when part of an eye is visible (fourth row), it exerts a strong constraint on the variation.}
	\label{fig:appendix_face_random}	
\end{figure*}

\subsection{Additional Results on Free-Form Mask Using Our \href{http://www.chuanxiaz.com/project/pluralistic/}{Interactive Demo}}

\begin{figure*}[h]
	\centering
	\includegraphics[width=\linewidth]{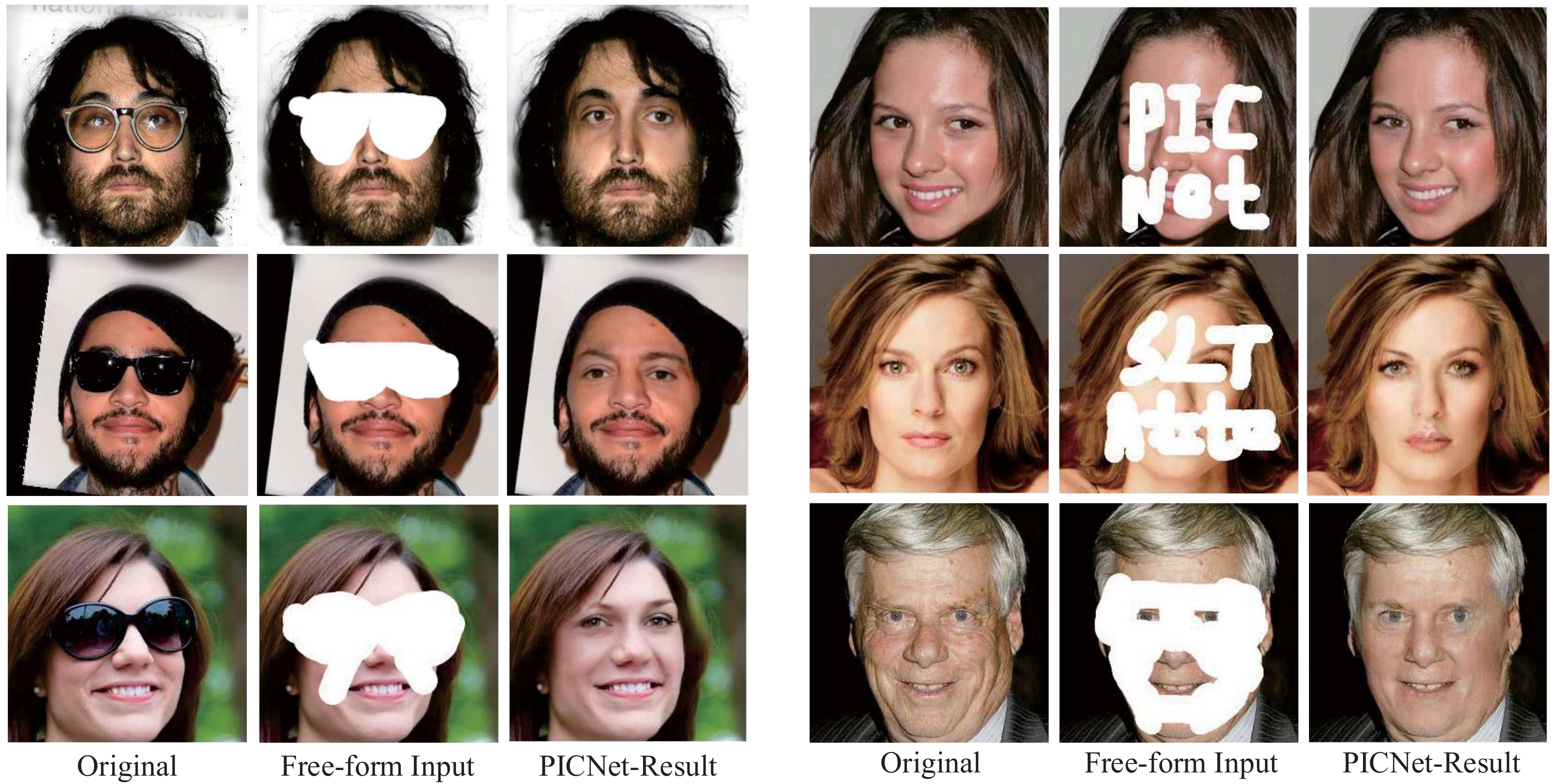}
	\caption{Face image editing results from our online interactive demo. The white mask regions will be normalized to gray mask as the input. It shows that our PICNet can be used to object removal and face editing.}
	\label{fig:appendix_face_playing}	
\end{figure*}

\subsection{Video for Additional Results}
Besides this document, we also included two video clips of additional results as part of the supplemental material. The first \href{https://youtu.be/tplhXSYX9us}{video}, shows free-from mask results on various datasets. The second \href{https://youtu.be/9V7rNoLVmSs}{video} consists of four parts to show multiple examples of center hole completion, random hole completion, comparison results with different training strategies and face editing of my self-portraits. 

\newpage
\section{Mathematical Derivation and Analysis}\label{math}
\subsection{Difficulties with Using the Classical CVAE for Image Completion}
\label{appsec:difficulties}

Here we elaborate on the difficulties encountered when using the classical CVAE formulation for pluralistic image completion, expanding on the shorter description in section~\ref{framework}.

\subsubsection{Background: Derivation of the Conditional Variational Auto-Encoder
	(CVAE)}

The broad CVAE framework of Sohn \etal\cite{sohn2015learning} is a straightforward conditioning of the classical VAE. Using the notation in our main paper, a latent variable $\mathbf{z}_{c}$ is assumed to stochastically generate the hidden partial image $\mathbf{I}_{c}$. When conditioned on the visible partial image $\mathbf{I}_{m}$, we get the conditional probability:
\begin{equation}
p(\mathbf{I}_{c}|\mathbf{I}_{m})=\int p_{\phi}(\mathbf{z}_{c}|\mathbf{I}_{m})p_{\theta}(\mathbf{I}_{c}|\mathbf{z}_{c},\mathbf{I}_{m})d\mathbf{z}_{c}
\end{equation}
The variance of the Monte Carlo estimate can be reduced by importance sampling to get
\begin{align}
p(\mathbf{I}_{c}|\mathbf{I}_{m}) & =\int q_{\psi}(\mathbf{z}_{c}|\mathbf{I}_{c},\mathbf{I}_{m})\frac{p_{\phi}(\mathbf{z}_{c}|\mathbf{I}_{m})}{q_{\psi}(\mathbf{z}_{c}|\mathbf{I}_{c},\mathbf{I}_{m})}p_{\theta}(\mathbf{I}_{c}|\mathbf{z}_{c},\mathbf{I}_{m})d\mathbf{z}_{c}\nonumber\\
& =\mathbb{E}_{\mathbf{z}_{c}\sim q_{\psi}(\mathbf{z}_{c}|\mathbf{I}_{c},\mathbf{I}_{m})}\left[\frac{p_{\phi}(\mathbf{z}_{c}|\mathbf{I}_{m})}{q_{\psi}(\mathbf{z}_{c}|\mathbf{I}_{c},\mathbf{I}_{m})}p_{\theta}(\mathbf{I}_{c}|\mathbf{z}_{c},\mathbf{I}_{m})\right]
\end{align}
Taking logs and apply Jensen's inequality leads to
\begin{align}
\log p(\mathbf{I}_{c}|\mathbf{I}_{m}) & \geq\mathbb{E}_{\mathbf{z}_{c}\sim q_{\psi}(\mathbf{z}_{c}|\mathbf{I}_{c},\mathbf{I}_{m})}\left[\log p_{\theta}(\mathbf{I}_{c}|\mathbf{z}_{c},\mathbf{I}_{m})-\log\frac{q_{\psi}(\mathbf{z}_{c}|\mathbf{I}_{c},\mathbf{I}_{m})}{p_{\phi}(\mathbf{z}_{c}|\mathbf{I}_{m})}\right]\nonumber\\
\mathcal{V} & =\mathbb{E}_{\mathbf{z}_{c}\sim q_{\psi}(\mathbf{z}_{c}|\mathbf{I}_{c},\mathbf{I}_{m})}\left[\log p_{\theta}(\mathbf{I}_{c}|\mathbf{z}_{c},\mathbf{I}_{m})\right]-\mathrm{KL}\left(q_{\psi}(\mathbf{z}_{c}|\mathbf{I}_{c},\mathbf{I}_{m})||p_{\phi}(\mathbf{z}_{c}|\mathbf{I}_{m})\right)
\label{eq:appendix_cvae}
\end{align}
The variational lower bound $\mathcal{V}$ totaled over all training data is jointly maximized \wrt the network parameters $\theta$, $\phi$ and $\psi$ in attempting to maximize the total log likelihood of the observed training instances.

\subsubsection{Single Instance Per Conditioning Label}

As is typically the case for image completion, there is only one training instance of $\mathbf{I}_{c}$ for each unique $\mathbf{I}_{m}$. This means that for the function $q_{\psi}(\mathbf{z}_{c}|\mathbf{I}_{c},\mathbf{I}_{m})$, $\mathbf{I}_{c}$ can simply be learnt into the network as a hardcoded dependency of the input $\mathbf{I}_{m}$, so $q_{\psi}(\mathbf{z}_{c}|\mathbf{I}_{c},\mathbf{I}_{m})\cong\hat{q}_{\psi}(\mathbf{z}_{c}|\mathbf{I}_{m})$. Assuming that the network for $p_{\phi}(\mathbf{z}_{c}|\mathbf{I}_{m})$ has similar or higher modeling power and there are no other explicit constraints imposed on it, then in training $p_{\phi}(\mathbf{z}_{c}|\mathbf{I}_{m})\rightarrow\hat{q}_{\psi}(\mathbf{z}_{c}|\mathbf{I}_{m})$, and the KL divergence in (\ref{eq:appendix_cvae}) goes to zero.

In this situation of zero KL divergence, we can rewrite the variational lower bound and replace $\hat{q}_{\psi}(\mathbf{z}_{c}|\mathbf{I}_{m})$ with $p_{\phi}(\mathbf{z}_{c}|\mathbf{I}_{m})$ without loss of generality, as
\begin{equation}
\mathcal{V}=\mathbb{E}_{\mathbf{z}_{c}\sim p_{\phi}(\mathbf{z}_{c}|\mathbf{I}_{m})}\left[\log p_{\theta}(\mathbf{I}_{c}|\mathbf{z}_{c},\mathbf{I}_{m})\right]
\end{equation}

\subsubsection{Unconstrained Learning of the Conditional Prior}

We can analyze how$\mathcal{V}$ can be maximized, by using Jensen's inequality again (reversing earlier use)
\begin{align}
\mathcal{V} & \leq\log\mathbb{E}_{\mathbf{z}_{c}\sim p_{\phi}(\mathbf{z}_{c}|\mathbf{I}_{m})}\left[p_{\theta}(\mathbf{I}_{c}|\mathbf{z}_{c},\mathbf{I}_{m})\right]\nonumber\\
& =\log\int p_{\phi}(\mathbf{z}_{c}|\mathbf{I}_{m})p_{\theta}(\mathbf{I}_{c}|\mathbf{z}_{c},\mathbf{I}_{m})d\mathbf{z}_{c}
\end{align}
By further applying H\"{o}lder's inequality (\ie $\left\Vert fg\right\Vert _{1}\leq\left\Vert f\right\Vert _{p}\left\Vert g\right\Vert _{q}$
for $\nicefrac{1}{p}+\nicefrac{1}{q}=1$), we get
\begin{align}
\mathcal{V} & \leq\log\left[\left|\int\left|p_{\phi}(\mathbf{z}_{c}|\mathbf{I}_{m})\right|d\mathbf{z}_{c}\right|\left|\int\left|p_{\theta}(\mathbf{I}_{c}|\mathbf{z}_{c},\mathbf{I}_{m})\right|^{\infty}d\mathbf{z}_{c}\right|^{\frac{1}{\infty}}\right]\quad(\text{by setting }p=1,q=\infty)\nonumber\\
& =\log\left[1\cdot\max_{\mathbf{z}_{c}}p_{\theta}(\mathbf{I}_{c}|\mathbf{z}_{c},\mathbf{I}_{m})\right]=\max_{\mathbf{z}_{c}}\log p_{\theta}(\mathbf{I}_{c}|\mathbf{z}_{c},\mathbf{I}_{m})
\end{align}
Assuming that there is a unique global maximum for $\log p_{\phi}(\mathbf{z}_{c}|\mathbf{I}_{m})$, the bound achieves equality when the conditional prior becomes a Dirac delta function centered at the maximum latent likelihood point
\begin{equation}
p_{\phi}(\mathbf{z}_{c}|\mathbf{I}_{m})\rightarrow\delta(\mathbf{z}_{c}-\mathbf{z}_{c}^{*})\quad\textrm{where }\mathbf{z}_{c}^{*}=\arg\max_{\mathbf{z}_{c}}\log p_{\theta}(\mathbf{I}_{c}|\mathbf{z}_{c},\mathbf{I}_{m})
\end{equation}
Intuitively, subject to the vagaries of stochastic gradient descent, the network for $p_{\phi}(\mathbf{z}_{c}|\mathbf{I}_{m})$ without further constraints will learn a narrow delta-like function that sifts out maximum latent likelihood value of $\log p_{\theta}(\mathbf{I}_{c}|\mathbf{z}_{c},\mathbf{I}_{m})$. 

As mentioned in section~\ref{framework}, although this narrow conditional prior may be helpful in estimating a single solution for $\mathbf{I}_{c}$ given $\mathbf{I}_{m}$ during testing during testing, this is poor for sampling a diversity of solutions. In our framework, the (unconditional) latent priors are imposed for the partial images themselves, which prevent this delta function degeneracy.

\subsubsection{CVAE with Fixed Prior}

An alternative CVAE variant~\cite{walker2016} assumes that conditional prior is independent of the $\mathbf{I}_m$ and fixed, so $p(\mathbf{z}_c|\mathbf{I}_m)\cong p(\mathbf{z}_c)$, where $p(\mathbf{z}_c)$ is a fixed distribution (\eg standard normal). This means
\begin{equation}
p(\mathbf{I}_c|\mathbf{I}_m) = \int p(\mathbf{I}_c|\mathbf{z}_c, \mathbf{I}_m) p(\mathbf{z}_c) d\mathbf{z}_c
\label{eq:cvae_fixed_prior}
\end{equation}
%An appropriate interpretation for this model is that $\mathbf{I}_m$ acts as a ``switch'' parameter to change between different likelihood functions $p(\mathbf{I}_c|\cdot, \mathbf{I}_m)$.
Now we can consider the case for a fixed $\mathbf{I}_m=\mathbf{I}^*_m$, and rewrite (\ref{eq:cvae_fixed_prior}) as
\begin{equation}
p_{\mathbf{I}^*_m}(\mathbf{I}_c) = \int p_{\mathbf{I}^*_m}(\mathbf{I}_c|\mathbf{z}_c) p(\mathbf{z}_c) d\mathbf{z}_c
\end{equation}
Doing so makes it obvious we can then derive the standard (unconditional) VAE formulation from here. Thus an appropriate interpretation of this CVAE variant is that it uses $\mathbf{I}_m$ as a ``switch'' parameeter to choose between different VAE models that are trained for the specific conditions.

Once again, this is fine if there are multiple training instances per conditional label. However, in the image completion problem, there is only one $\mathbf{I}_c$ per unique $\mathbf{I}_m$, so the condition-specific VAE model will simply ignore the sampling ``noise'' and learn to predict the single instance of $\mathbf{I}_c$ from $\mathbf{I}_m$ directly, \ie $p(\mathbf{I}_c|\mathbf{z}_c, \mathbf{I}_m) \approx p(\mathbf{I}_c|\mathbf{I}_m)$, which incidentally achieves equality for the variational lower bound. This results in negligible variation of output despite now sampling from $p(\mathbf{z}_c)=\mathcal{N}(0,1)$.

Our framework resolves this in part by defining all (unconditional) partial images of $\mathbf{I}_c$ as sharing a common latent space with adaptive priors, with the likelihood parameters learned as an unconditional VAE, and further coupling on the conditional portion (\ie the generative path) to get a more distinct but regularized estimate for $p(\mathbf{z}_c|\mathbf{I}_m)$.

\subsection{Joint Maximization of Unconditional and Conditional Variational Lower
	Bounds}
\label{appsec:joint_maximization}

The overall training loss function (\ref{eq:total_loss}) used in our framework has a direct link to jointly maximizing the unconditional and unconditional variational lower bounds, respectively expressed by (\ref{eq:VAE}) and (\ref{eq:mixed_models}). Using simplified notation, we rewrite these bounds respectively as:
\begin{align}
\mathcal{B}_{1} & =\mathbb{E}_{q_{\psi}}\log p_{\theta}^{r}-\mathrm{KL}(q_{\psi}||p_{z_{c}})\nonumber\\
\mathcal{B}_{2} & =\lambda\left(\mathbb{E}_{q_{\psi}}\log p_{\theta}^{r}-\mathrm{KL}(q_{\psi}||p_{z_{c}})\right)+\mathbb{E}_{p_{\phi}}\log p_{\theta}^{g}
\end{align}
To clarify, $\mathcal{B}_{1}$ is the lower bound related to the unconditional log likelihood of observing $\mathbf{I}_{c}$, while $\mathcal{B}_{2}$ relates to the log likelihood of observing $\mathbf{I}_{c}$ conditioned on $\mathbf{I}_{m}$. The expression of $\mathcal{B}_{2}$ reflects a blend of conditional likelihood formulations with and without the use of importance sampling, which are matched to different likelihood models, as explained in section~\ref{sec:rec_vs_gen}. Note that the $(1-\lambda)$ coefficient from (\ref{eq:mixed_models}) is left out here for simplicity, but there is no loss of generality since we can ignore a constant factor of the true lower bound if we are simply maximizing it.

We can then define a combined objective function as our maximization goal
\begin{align}
\mathcal{B}&= \beta \, \mathcal{B}_{1}+\mathcal{B}_{2}\nonumber\\
&= (\beta+\lambda)\mathbb{E}_{q_{\psi}}\log p_{\theta}^{r}+\mathbb{E}_{p_{\phi}}\log p_{\theta}^{g}-\left[\beta\mathrm{KL}(q_{\psi}||p_{z_{c}})+\lambda\mathrm{KL}(q_{\psi}||p_{\phi})\right]
\label{eq:total_bound}
\end{align}
with $\beta\geq0$.

To understand the relation between $\mathcal{B}$ in (\ref{eq:total_bound}) and $\mathcal{L}$ in (\ref{eq:total_loss}), we consider the equivalence of:
\begin{equation}
-\mathcal{B}\cong\mathcal{L}=\alpha_{\mathrm{KL}}(\mathcal{L}_{\mathrm{KL}}^{r}+\mathcal{L}_{\mathrm{KL}}^{g})+\alpha_{\mathrm{app}}(\mathcal{L}_{\mathrm{app}}^{r}+\mathcal{L}_{\mathrm{app}}^{g})+\alpha_{\mathrm{ad}}(\mathcal{L}_{\mathrm{ad}}^{r}+\mathcal{L}_{\mathrm{ad}}^{g})
\end{equation}
Comparing terms
\begin{equation}
\mathcal{L}_{\mathrm{KL}}^{r}\cong\mathrm{KL}(q_{\psi}||p_{z_{c}}),\quad\mathcal{L}_{\mathrm{KL}}^{g}\cong\mathrm{KL}(q_{\psi}||p_{\phi})\quad\Rightarrow\beta=\lambda=\alpha_{\mathrm{KL}}
\end{equation}

For the reconstructive path that involves sampling from the (posterior) importance function $q_{\psi}(\mathbf{z}_{c}|\mathbf{I}_{c})$ of (\ref{eq:CVAE_with_prior}), we can substitute $(\beta+\lambda)=2\alpha_{\mathrm{KL}}$ and get the reconstructive log likelihood formulation as
\begin{equation}
-\mathbb{E}_{q_{\psi}}\log p_{\theta}^{r}\cong\frac{\alpha_{\mathrm{app}}}{2\alpha_{\mathrm{KL}}}\mathcal{L}_{\mathrm{app}}^{r}+\frac{\alpha_{\mathrm{ad}}}{2\alpha_{\mathrm{KL}}}\mathcal{L}_{\mathrm{ad}}^{r}
\end{equation}
Here, $\mathbf{I}_{c}$ is available, with $\mathcal{L}_{\mathrm{app}}^{r}$ reconstructing both $\mathbf{I}_{c}$ and $\mathbf{I}_{m}$ as in (\ref{eq:app_loss_rec}), while $\mathcal{L}_{\mathrm{ad}}^{r}$ involves GAN-based pairwise feature matching (\ref{eq:ad_loss_rec}).

For the generative path that involves sampling from the conditional prior $p_{\phi}(\mathbf{z}_{c}|\mathbf{I}_{m})$, we have the generative log likelihood formulation as
\begin{equation}
-\mathbb{E}_{p_{\phi}}\log p_{\theta}^{g}\cong\alpha_{\mathrm{app}}\mathcal{L}_{\mathrm{app}}^{g}+\alpha_{\mathrm{ad}}\mathcal{L}_{\mathrm{ad}}^{g}
\end{equation}
As explained in sections~\ref{sec:rec_vs_gen} and \ref{sec:dual_pipeline_structure}, the generative path does not have direct access to $\mathbf{I}_{c}$, and this is reflected in the likelihood $p_{\theta}^{g}$ in which the instances of $\mathbf{I}_{c}$ are ignored. Thus $\mathcal{L}_{\mathrm{app}}^{g}$ is only for reconstructing $\mathbf{I}_{m}$ in a deterministic auto-encoder fashion as per (\ref{eq:app_loss_gen}), while $\mathcal{L}_{\mathrm{ad}}^{g}$ in (\ref{eq:ad_loss_gen}) only tries to enforce that the generated distribution be consistent with the training set distribution (hence without per-instance knowledge), as implemented in the form of a GAN.

\newpage
\section{Architectural Details}\label{architecture} 

\begin{figure*}[h]
	\centering
	\includegraphics[width=0.98\linewidth]{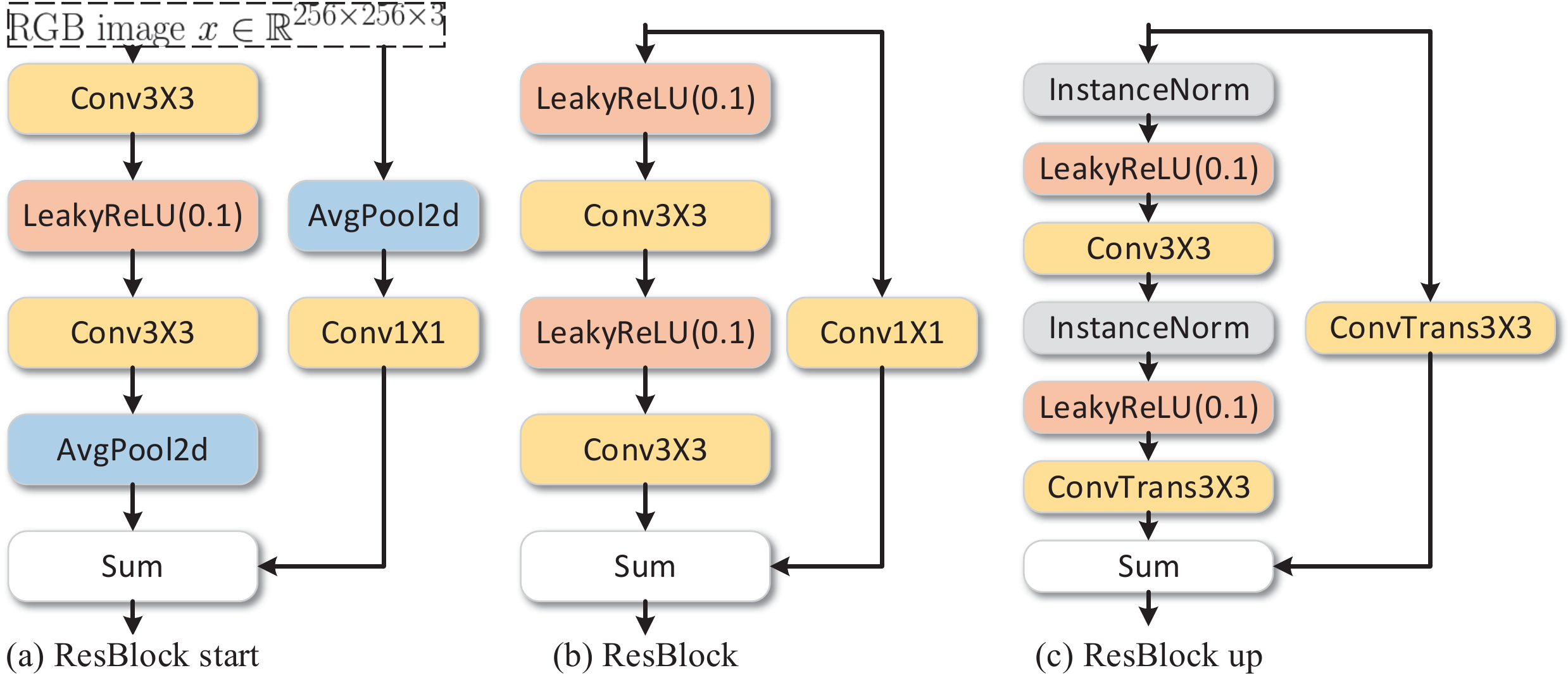}
	\caption{Illustration of the Residual Block used in our model. (a) The starter Residual Block for the encoder (representation) and discriminator networks. (b) A Residual Block in the encoder (representation), inference and discriminator networks. (c) A Residual Block in the decoder (generator) network.}
	\label{fig:appendix_structure}	
\end{figure*}

Our {\bf pluralistic image completion} network (\textbf{PICNet}) architecture is inspired by SA-GAN \cite{zhang2018self} and BigGAN, but features several important modifications that enable us to train for this image-conditional generation task. We first replace the batch normalization with instance normalization in the generation network ({\bf ResBlock up} in Fig.~\ref{fig:appendix_structure}), and remove the batch normalization in our other networks,  (\ie. the representation, inference and discriminator networks comprising {\bf ResBlock start} and {\bf ResBlock} in Fig.~\ref{fig:appendix_structure}), because different holes will affect the means and variances in each batch. {\bf ResBlock down} is similar to {\bf ResBlock}, in which we add the average pooling layer after Conv$3\times3$ and Conv$1\times1$.

\begin{table*}[h]
	
	\begin{minipage}{0.30\textwidth}
		\centering
		\includegraphics[width=\linewidth]{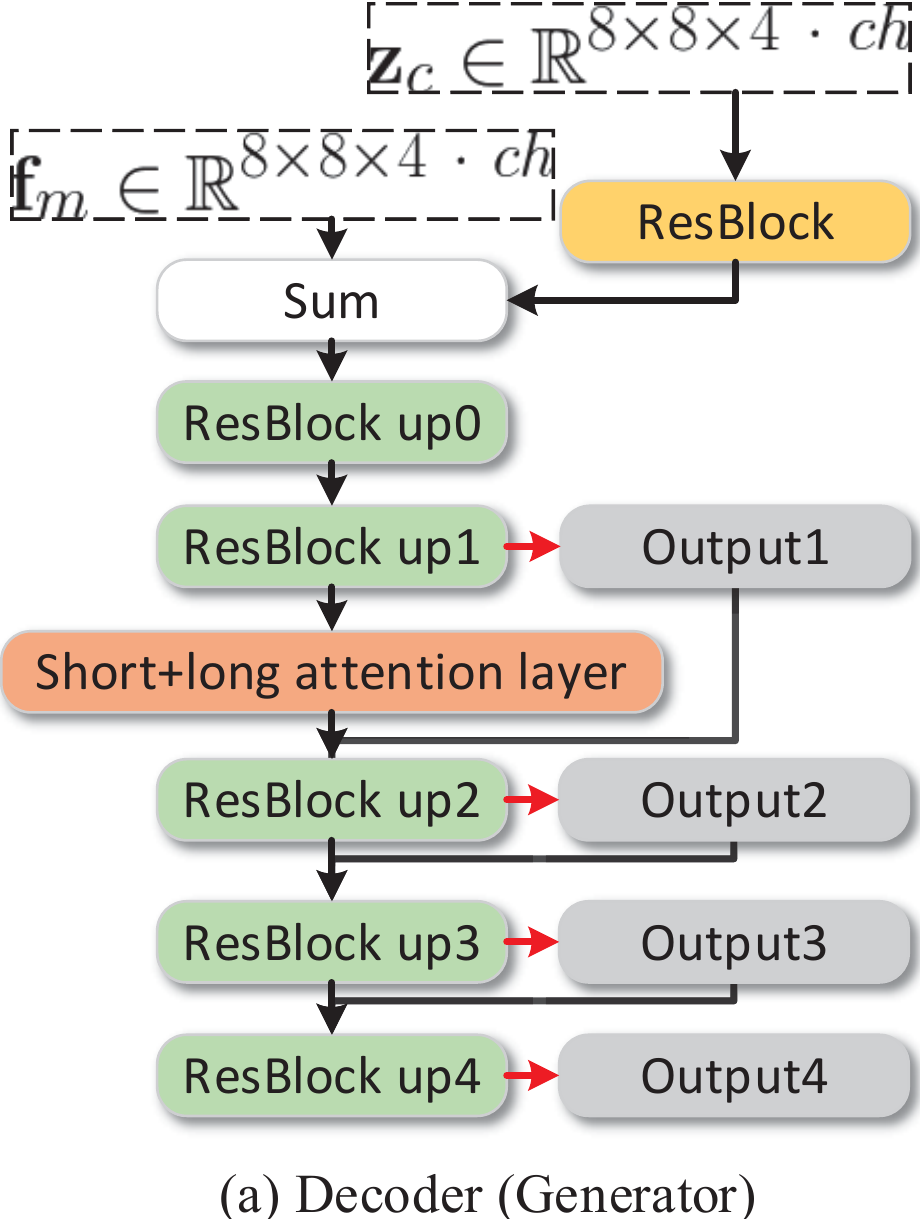}
		\label{fig:appendix_generator}
	\end{minipage}	
	\hfill
	\begin{minipage}{0.28\textwidth}
		\begin{center}
			\renewcommand{\arraystretch}{1.5}
			\begin{tabular}{c}
				\toprule[2pt]
				RGB image $x\in\mathbb{R}^{256\times256\times3}$ \\
				\hline
				ResBlock start $128\times128\times1~\cdot~ch$ \\
				\hline
				ResBlock down $64\times64\times2~\cdot~ch$ \\
				\hline
				ResBlock down $32\times32\times4~\cdot~ch$ \\
				\hline
				ResBlock down $16\times16\times4~\cdot~ch$ \\
				\hline
				ResBlock down $8\times8\times4~\cdot~ch$ \\
				\bottomrule[2pt]
				(b) Encoder (Representation)
			\end{tabular}
		\end{center}
	\end{minipage}
	\hfill
	\begin{minipage}{0.30\textwidth}
		\begin{center}
			\renewcommand{\arraystretch}{1.5}
			\begin{tabular}{c}
				\toprule[2pt]
				RGB image $x\in\mathbb{R}^{256\times256\times3}$ \\
				\hline
				ResBlock start $128\times128\times1~\cdot~ch$ \\
				\hline
				ResBlock down $64\times64\times2~\cdot~ch$ \\
				\hline
				ResBlock down $32\times32\times4~\cdot~ch$ \\
				\hline
				Self-Attention Layer $32\times32\times4~\cdot~ch$ \\
				\hline
				ResBlock down $16\times16\times4~\cdot~ch$ \\
				\hline
				ResBlock down $8\times8\times4~\cdot~ch$ \\
				\hline
				ResBlock $7\times7\times4~\cdot~ch$ \\
				\hline
				LeakyReLU(0.1), Conv, $6\times6\times1$ \\
				\bottomrule[2pt]
				(c) Discriminator
			\end{tabular}
		\end{center}
	\end{minipage}
	\caption{Architectures for our framework, where $ch$ represents the base channel width. For the output layer, we use the LeakyReLU(0.1), Conv$3\times3$ and Tanh at all scales.}
	\label{table:architecture}
\end{table*}

The \textbf{Infer1} network only consists of one Residual Block, for self-inferring the latent distribution of the ground truth $\mathbf{I}_c$ (treated as known in the reconstructive path), while the \textbf{Infer2} network consists of seven Residual Blocks, which are applied to predict the latent distribution of $\mathbf{I}_c$ (treated as unknown in the generative path) based on the visible pixels $\mathbf{I}_m$. 

\section{Experimental Details}\label{experiment}

Our network is implemented in Pytorch v0.4.0, and employs the architectures of Appendix~\ref{architecture}. To reduce memory cost, we restrained the feature channel width to $4~\cdot~ch$ and selected $ch=32$. We experimented with different channels with largest being $16~\cdot~ch=1024$, but found that the improvement was not obvious. In addition, we applied the self-attention layer of the discriminator and the short+long term attention layer of the generator on a $32\times32$ feature size. Spectral Normalization is used in all networks. All networks are initialized with Orthogonal Initialization and trained from scratch with a fixed learning rate of $\lambda=10^{-4}$. We used the Adam optimizer with $\beta_1=0$ and $\beta_2=0.999$. 

The final weights we used were $\alpha_\text{KL}=\alpha_\text{app}$=20, $\alpha_\text{ad}$=1. The KL loss and appearance matching loss weights come from the variational \emph{lower bound}. Since the appearance matching loss is used in four output scales, the final weight for the KL loss is $\alpha_\text{KL}=\alpha_\text{KL}\times N_\text{scale}$, where $N_\text{scale}$ is the number of output scales. We also tried different values of $\alpha_\text{KL}$ and $\alpha_\text{app}$, and found that the bigger the KL loss weight, the greater the diversity of the generated $\mathbf{I}_c^{'}$, but it was also harder to retain the appearance consistency of the generated $\mathbf{I}_c^{'}$ to the visible region $\mathbf{I}_m$. The values of $\alpha_\text{app}$ and $\alpha_\text{ad}$ were obtained from $\alpha$-GAN. We experimented with the number of $D$ steps per $G$ step (varying it from 1 to 5), and found that one $D$ step per $G$ step gave the best results. When $\alpha_\text{app}$ is smaller than 1, we can use two or four $D$ steps per $G$ step, but the full generated $\mathbf{I}_g^{'}$ does not reconstruct the original conditional visible regions $\mathbf{I}_m$ well. When $\alpha_\text{app}$ is larger than 100, we needed two or four $G$ steps per $D$ step, if not the discriminator loss will become zero and the generated $\mathbf{I}_c^{'}$ will be blurry.

We trained each model on a single GPU, with a batch size of 20 on a GTX 1080TI (11GB) and 32 on a NVIDIA V100 (16GB). Training models for centered holes of Paris and CelebA-HQ takes roughly 3 days, while for ImageNet and Places2 it takes roughly 2 weeks. On the other hand, training models for random irregular and un-centered holes takes about twice the time compared to models for centered holes. Moreover, since the prior distribution of random holes $p(\mathbf{z})=\mathcal{N}_m(\mathbf{0},\sigma^2(n)\mathbf{I})$ is changed with the number of pixels in each hole $n$, the training loss may sometimes change abruptly due to the KL loss component. 

\end{document}